\pdfoutput=1

\documentclass[11pt]{article}

\usepackage[final]{acl}

\usepackage{times}
\usepackage{latexsym}
\usepackage{booktabs}
\usepackage{hhline}
\usepackage{multirow}
\usepackage{hyperref}
\usepackage{soul}
\usepackage{tabularx}
\usepackage{caption}
\usepackage{subcaption}
\usepackage{adjustbox}
\usepackage{arydshln}
\usepackage{makecell}
\usepackage{array}
\usepackage{enumitem}
\usepackage{stfloats}
\usepackage{float}
\usepackage{placeins}

\usepackage[T1]{fontenc}

\usepackage[utf8]{inputenc}
\usepackage{csquotes}

\usepackage{microtype}

\usepackage{inconsolata}

\usepackage{graphicx}

\newcommand{\ls}{German Easy language}

\title{Images Speak Volumes: User-Centric Assessment of Image Generation for Accessible Communication}

 \author{Miriam Ansch\"{u}tz \and Tringa Sylaj \and Georg Groh \\
         School for Computation, Information and Technology \\
         Technical University of Munich, Germany\\
         \texttt{\{\href{mailto:miriam.anschuetz@tum.de}{miriam.anschuetz}, \href{mailto:tringa.sylaj@tum.de}{tringa.sylaj}\}@tum.de}, \texttt{grohg@in.tum.de}
         }

\begin{document}
\maketitle
\begin{abstract}
Explanatory images play a pivotal role in accessible and easy-to-read (E2R) texts. However, the images available in online databases are not tailored toward the respective texts, and the creation of customized images is expensive. In this large-scale study, we investigated whether text-to-image generation models can close this gap by providing customizable images quickly and easily. We benchmarked seven, four open- and three closed-source, image generation models and provide an extensive evaluation of the resulting images. In addition, we performed a user study with people from the E2R target group to examine whether the images met their requirements. We find that some of the models show remarkable performance, but none of the models are ready to be used at a larger scale without human supervision. Our research is an important step toward facilitating the creation of accessible information for E2R creators and tailoring accessible images to the target group's needs.
\end{abstract}

\section{Introduction}
Easy-to-read (E2R) and its German derivative \textit{Leichte Sprache} (Easy Language) are accessibility- and readability-enhanced versions of language. They follow a strict ruleset and are targeted at people with disabilities, learning difficulties, or low literacy \citep{din-spec-ls}. The creation of a more accessible version of an original text is called text simplification (TS). Since this process is laborsome, previous work explored the applicability of large language models to facilitate or even automatize the creation of E2R texts \citep{madina-e2r}. For \ls{}, \citet{schomacker-accessible} investigated how well the currently available, text-oriented models and datasets comply with the ruleset of \ls{} and multiple open-source automatic TS models for German exist \citep{anschutz-german-simp, stodden-deplain}.

\begin{figure}[ht]
    \centering
    \includegraphics[width=\linewidth]{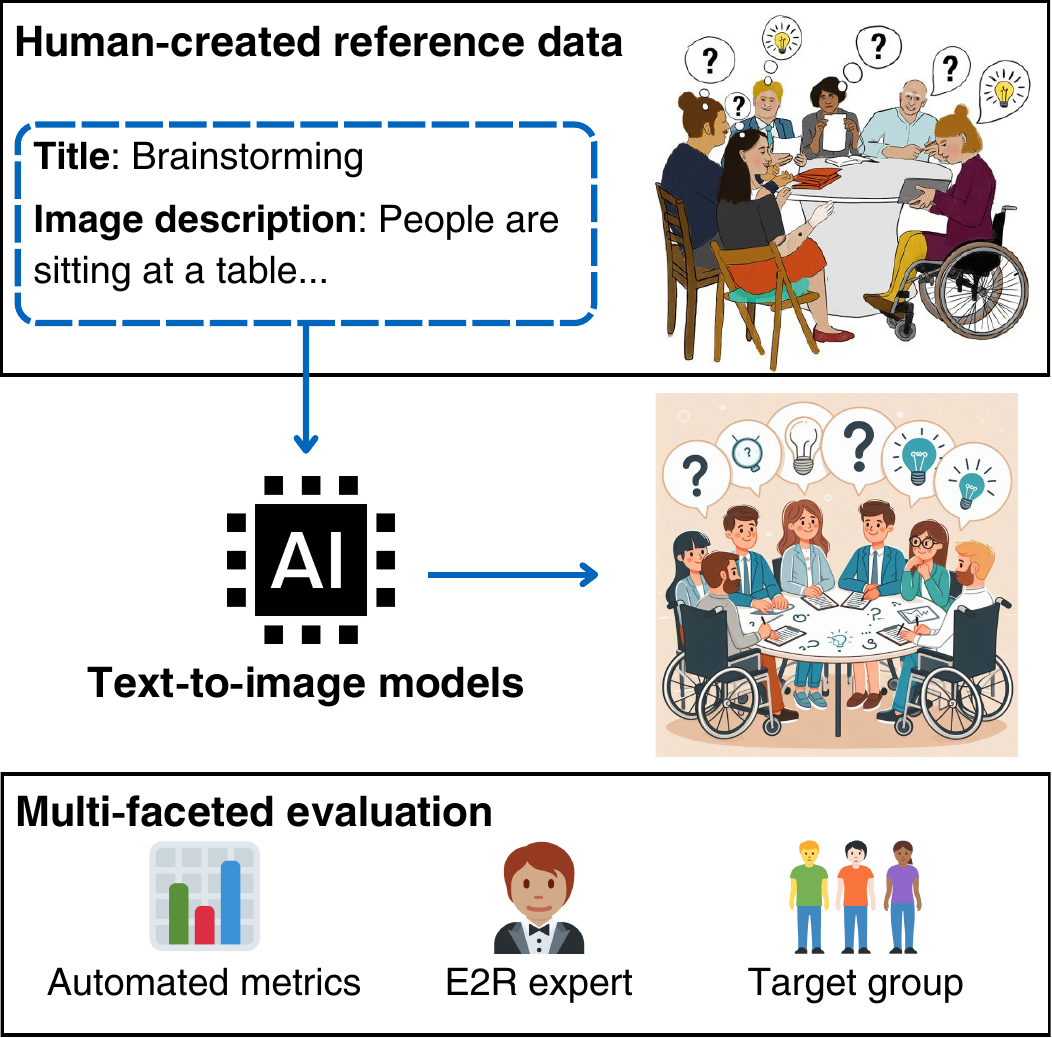}
    \caption{Overview of our approach: We selected a human-created reference dataset that was validated by the target group already. Based on the images' titles and descriptions, we used seven different text-to-image models to recreate the original images. Then, we evaluated the generated images across multiple aspects using automated and human evaluation.\\ © JSCHKA Kommunikationsdesign | \href{www.jschka.de}{www.jschka.de}}
    \label{fig:vis_abs}
\end{figure}


However, one important feature of E2R texts is that they are illustrated with images that improve and facilitate the text's understanding even further. The guidelines in the DIN-SPEC 33429 \citep{din-spec-ls} recommend that these images should be created specifically for each text and that they should be up-to-date and close to the target group's everyday life. 
Even though large image databases exist\footnote{e.g. \url{https://www.lag-sb-rlp.de/projekte/bildergalerie-leichte-sprache}} that were reviewed and validated by the target group, these images were created for a general purpose and cannot be altered by the text creators. In addition, while human artists are unchallenged in creating the most targeted and realistic images, their employment is financially infeasible for most E2R translators. Therefore, our work explores whether text-to-image (T2I) models can solve this problem by creating quickly available, flexible, and cheap images. An overview of our approach is presented in \autoref{fig:vis_abs}.\\
Our contribution can be summarized as follows:
\begin{itemize}
    \item We benchmark seven text-to-image models (four open-sourced and three closed-source) on their ability to create images for accessible communication.
    \item The resulting images are published as a dataset consisting of 2,217 images\footnote{\href{https://github.com/MiriUll/Image-Generation-for-Accessible-Communication}{https://github.com/MiriUll/Image-Generation-for-Accessible-Communication}}. This dataset is relevant for text creators searching for a large and diverse image database as well as for AI researchers who want to train models or evaluation metrics for this task.
    \item We manually reviewed 560 images and annotated them by their closeness to the prompt, correctness, bias toward people with disabilities, and suitability for the target group. Our findings indicate that the quality of the generated images is highly dependent on the depicted content and the T2I models used and that even the best models cannot be utilized for a broader scale without further restrictions.
    \item We conducted a user study with seven people from the target group and report their opinions and preferences about the generated images.
\end{itemize}

\section{Related work}
Previous work has utilized images to enhance text accessibility, particularly in fields such as language learning. \citet{geislinger-l2learners-eyetracking} developed an iPad app for language learners that included an eye-tracking feature. When a reader focused on a word for a longer time, they retrieved a picture illustrating that word and showed it next to the text to improve the text's understanding. Similarly, \citet{singh-textbook-images} and \citet{schneider-retrieval-textbook-benchmark} focused on retrieving images for textbooks to improve the learning experience and make the books more appealing. To train and benchmark those retrieval models, \citet{wang-motif-complex-words} published the MOTIF dataset. The dataset consists of sentences with complex words within the sentences and images that represent the context of the sentences. The complex word is highlighted within the images to give an easy-to-grasp explanation of those complex words. However, all of the previous methods only search for images in existing image databases and explore the capabilities of image retrieval methods. 
In contrast to this, our focus lies on the generation of new images and benchmarking models to create those images. A similar task was proposed by \citet{kiesel-clef-touche}, who tried to strengthen argumentation chains by providing images supporting the argument's premise. Nevertheless, to the best of our knowledge, this is the first work to explore image generation to automatically enhance accessible communication.

There exist multiple studies about the characteristics of image generation models, but none of them addresses their applicability to accessible communication.
\citet{mack-bias-t2i} benchmarked different T2I models like DALL-E 2, Stable Diffusion, and Midjourney about how they depict disabilities. Even though the prompts described different forms of disabilities, the models mostly depicted disabled people as sitting in wheelchairs. The findings were repeated in the study from \citet{tevissen-disability-study}. The author investigated the latest Stable Diffusion checkpoints, SDXL and Stable Diffusion 3, DALL-E 3, and Midjourney. Again, people with disabilities were depicted very stereotypically: as old and sad people sitting in wheelchairs.

In our study, we include people from the target group and also report their perspectives on image generation models for accessible communication. Similar user studies were conducted in previous work. \citet{huh-GenAssist} aim to make image generation as a process more accessible. They created a framework called GenAssist, in which blind and low-vision creators can ask questions about the image to determine whether the image generation models followed their prompts or whether additional content was added. A user study with the target group proved that the tool made visual creations more accessible.
Another target group study was conducted by \citet{edwards-study-image-description}, who worked with people with disabilities and asked them how their disability should be depicted in generated images. They especially focused on disability descriptions and the best level of detail for these descriptions. Similarly, \citet{das-alt-text-study} worked together with image creators and screen reader users to evaluate images' alt texts from different perspectives. They report that manually created alt texts are often too subjective and that prompts for T2I models cannot be used as alt text alternatives.

\section{Methodology}
Our study investigates whether the latest T2I models can create images suitable for E2R texts. For this, we use an open-source database of images for \ls{} and try to recreate the images based on their title and descriptions. 
Most of the images in E2R image databases are cartoon images since they are often easier than photo-realistic images, and the readers don't get confused by actual people. Therefore, we only focus on the generation of cartoon images as well.
\subsection{Reference dataset}

\begin{figure}[ht]
    \centering
    \includegraphics[width=\linewidth]{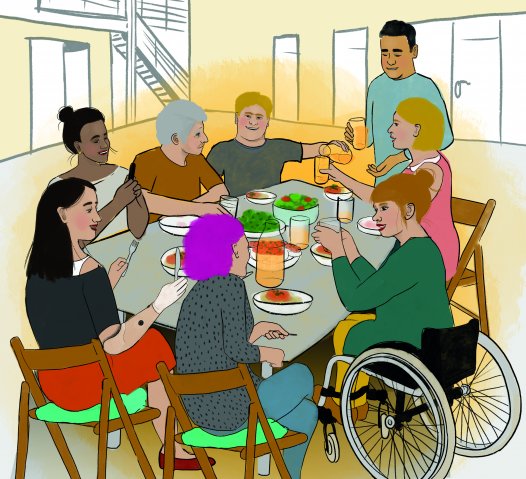}
    \caption{Example image for the word \enquote{Inclusion} from the Leichte Sprache image gallery. The image description is \enquote{A group of very different people \- with and without disabilities \- is sitting at a table and eating together.}\\ © JSCHKA Kommunikationsdesign | \href{www.jschka.de}{www.jschka.de}}
    \label{fig:inclusion_example}
\end{figure}

Our target dataset is the publicly available Leichte Sprache image gallery\footnote{\url{https://www.lag-sb-rlp.de/projekte/bildergalerie-leichte-sprache}} from the LAG Selbsthilfe von Menschen mit Behinderungen und chronischen Erkrankungen Rheinland-Pfalz e.V. (State working group for self-help for people with disabilities and chronic illnesses Rhineland-Palatinate, Germany), a state-level organization uniting self-help associations and groups of individuals
with disabilities or chronic illnesses and their relatives. It offers 413 images within 16 categories drawn by the artist Juliane Kriegereit\footnote{JSCHKA Kommunikationsdesign | \href{www.jschka.de}{www.jschka.de}}. The images were created for E2R texts and reviewed by the target group. An example image is shown in \autoref{fig:inclusion_example}. The images' license is very permissive to enable content creators to illustrate their texts. The categories are targeted to people with disabilities and cover areas like assisting technologies, diseases, and body parts. Each picture comes with a topic that is depicted and a description of the image's contents. 

For our experiments, we randomly selected five images per category, yielding a dataset of $16 \times 5 = 80$ reference images in total. We translated the image titles and descriptions into English using ChatGPT \citep{openai-gpt4} since some image generation models only work with English prompts.

\subsection{Text2Image models}

Our model selection featured a mix of open and closed-source models, SOTA and older models, as well as models of various sizes, culminating in a comprehensive evaluation of seven models in total. An overview of the models can be found in \autoref{tab:models_limit} in the Appendix.

We constructed the model prompts as \enquote{Cartoon picture of \{title\} - \{description\}} where we filled the placeholders with the values from the dataset and used the same prompts for all models.

For the open-sourced models, we utilized various versions of Stable Diffusion \citep{rombach2022highresolution} and Würstchen \citep{pernias2023wuerstchen}. 
Stable Diffusion v1.4, v2.1 base, and v3 were employed to generate 512x512 pixel images. 
For SD3, we used the default parameters optimized for the output quality: \textit{num\_inference\_steps} was configured to 28, defining the number of denoising steps the model takes during image generation. A higher number of inference steps generally leads to finer details and improved image quality. Additionally, the \textit{guidance\_scale} was set to 7.0, indicating the strength of the conditioning on the input text prompt. A higher guidance scale helps produce images that are more closely aligned with the given text descriptions, ensuring the semantic accuracy of the generated images.

Würstchen \citep{pernias2023wuerstchen} is another diffusion model where the text-conditional component functions within a significantly compressed latent space of images, attaining a 42x spatial compression. This enables the model to be much more time- and memory-efficient, significantly reducing training and inference time. Würstchen was used to produce higher-resolution images at 1024x1024 pixels, using a \textit{prior\_guidance\_scale} set to 4.0, which similarly influences the model's adherence to the textual input.

For the closed-sourced model we focused on DALL-E-3 \citep{ramesh2021zeroshot}, Midjourney \footnote{\url{https://www.midjourney.com/} Last accessed: Jul 2024}, and Artbreeder \footnote{\url{https://www.artbreeder.com/} Last accessed: Jul 2024}. We accessed DALL-E-3 through the Bing Image Creator by Microsoft \footnote{\url{https://www.bing.com/images/create} Last accessed: Jul 2024}, which Microsoft states is powered by an advanced version of the DALL-E model. We used the free version, which allows 15 prompts per day. For Artbreeder, we use the Composer model, which is a GAN architecture \citep{goodfellow2014generativeadversarialnetworks} incorporating elements of BigGAN \citep{brock2019largescalegantraining} and StyleGAN \citep{karras2019stylebasedgeneratorarchitecturegenerative}.



\subsection{Evaluation}
We evaluated our generated images on different aspects, which include the closeness to the reference images, how well the models follow the image description in the prompt, and the image correctness.

The most popular automatic evaluation metric to measure the quality of a generated image is the Inception Score \citep{Salimans-Inception-score}. However, it compares the generated images against photo-realistic reference images from the CIFAR-10 dataset that are limited in the items they depict. Hence, the inception score is not suitable for our cartoon-style images \citep{provenbessel-ComicGAN, barratt-inception-critique}. To automatically assess the quality of our generated images, we used the Fréchet Inception Distance (FID). In contrast to the inception score, FID compares the generated images against a set of user-selected reference images. It estimates the distributions of the reference and generated image sets and reports the distance between the two distributions. Therefore, a lower FID score indicates better matches with the reference images and, thus, a better overall image quality. For our experiments, we used the FID implementation by PyTorch Lightning\footnote{\url{https://lightning.ai/docs/torchmetrics/stable/image/frechet_inception_distance.html}}.

The second aspect of our evaluation is how well the generated images follow the image descriptions. For this, we evaluate two different metrics. The first metric is Contrastive Language-Image Pre-training (CLIP), which is trained to determine if an image and a text are paired together \citep{radford-CLIP}. It encodes the images and texts into a joint embedding space and selects the most probable pairs among them. For our experiments, we use the pre-trained CLIP ViT-L/14@336px model that achieves the highest accuracies according to the authors (Figure 10 in \citet{radford-CLIP}). 

Our third metric, TIFA \cite{hu-tifa}, also evaluates the fit between an image and its description, similar to the CLIP score, but chooses a different approach: visual question answering. For this, \citet{hu-tifa} created a three-step pipeline: First, an LLM creates single-choice, multiple-choice, and free-form questions and their answers from the image descriptions. Each question is categorized by the elements it is asking for, e.g., color or location, and the number of questions and the element types vary among the different images. Then, a second question-answering model tries to answer the questions based on the image descriptions. Only questions that receive the same answers from both systems are kept for visual evaluation. Finally, a visual question-answering model answers the questions by looking at the images. The image-based accuracy of the answers indicates how faithful the image is to the image description. TIFA incorporates the accuracy metric, and thus, the scores range between 0 and 1. The authors show that TIFA has a much higher correlation with human judgments than previous metrics like CLIP \citep{hessel-clip}.

For our study, we use the pre-trained checkpoints for the different parts of the pipeline. For the question generation, we use the author's fine-tuned Llama2 \citep{touvron-llama2} model. For the question filtering, we use a UnifiedQA \citep{khashabi-unifiedqa} model. The set of questions was only created once per image prompt and then used for all model evaluations. With this, we reduce biases in the scores that could come from non-determinism in the question generation or filtering models. Finally, for the visual question-answering, the authors compared different models. We selected the model with the highest correlation with human judgment, according to the authors, which is mPLUG-large \citep{li-mplug}. 

\section{Results}
To obtain a diverse image collection, we created up to four images per model an prompt. We investigate seven different models, and thus, expected to generate $80 \times 4 \times 7 = 2,240$ images.
However, we only obtained 2,217 images in total. For some descriptions, Copilot's DALL-E interface returned less than four images, resulting in only 297 instead of 320 images from DALL-E. In addition, Stable Diffusion 1 and DALL-E marked some of our image descriptions as offensive and blocked the input. For Stable Diffusion, this resulted in images that were all black. We followed this approach and added four black images for each of the five blocked inputs for DALL-E as well. This results in 51 images that are blackened.
\subsection{Automatic evaluation}
To further assess our generated images, we calculated metric scores as shown in  \autoref{tab:tifa_clip_scores}. The best FID scores are achieved by Stable Diffusion 3 and Midjourney. This indicates that their style of images comes closest to the style of the reference image and that the images have similar features.
\begin{figure*}
    \centering
    \includegraphics[width=\textwidth]{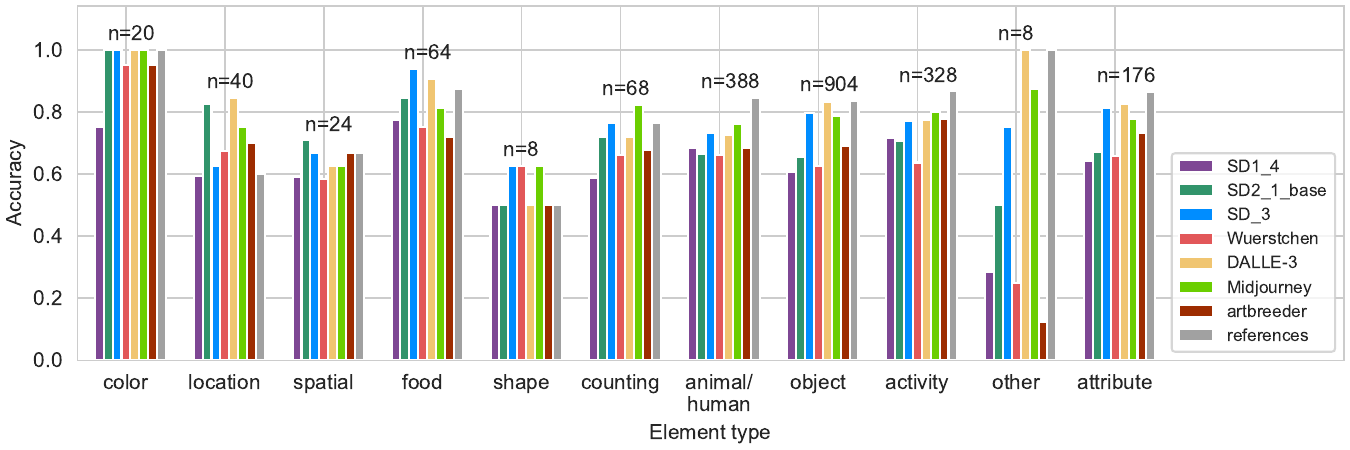}
    \caption{TIFA accuracies for different element types targeted by the TIFA questions. The performance of the models depends on the content that they have to depict. Images with all black content were filtered.}
    \label{fig:tifa_elements}
\end{figure*}

\begin{table}[ht]
    \centering
    \begin{tabular}{lccc}\toprule
        \textbf{Model} & \textbf{FID}$\downarrow$ & \textbf{CLIP}$\uparrow$ & \textbf{TIFA}$\uparrow$ \\
        \midrule
        \textbf{SD1\_4} & 1.49 & 0.22 & 0.58 \\
        \textbf{SD2\_1\_base} & 1.37 & 0.24 & 0.68 \\ 
        \textbf{SD\_3} & \textbf{0.89} &\textbf{ 0.27} &\textbf{ 0.78} \\
        \textbf{W\"urstchen} & 0.90 & 0.24 & 0.65 \\
        \midrule
        \textbf{DALL-E-3} & 1.26 & \textbf{0.26} & 0.74 \\
        \textbf{Midjourney} & \textbf{0.90} & \textbf{0.26} & \textbf{0.78} \\
        \textbf{Artbreeder} & 1.52 & 0.25 & 0.70 \\
        \midrule
        \textbf{References}& - & 0.27 & 0.84 \\
        \bottomrule
    \end{tabular}
    \caption{Macro-averaged automatic evaluation scores to evaluate the images' distribution compared to the references (FID) and their closeness to the prompts (CLIP and TIFA). Stable Diffusion 3, DALL-E-3, and Midjourney come closest to the human-created reference images.}
    \label{tab:tifa_clip_scores}
\end{table}

The CLIP and TIFA metrics evaluate how well the images align with the image descriptions. These metrics don't rely on the reference images, and hence, we calculated the scores on the references as well. We manually set the scores to 0 for all-black images with blocked contents and ignored one image title in the CLIP evaluation whose prompt was too long for the CLIP model. For both metrics, Stable Diffusion 3 has the highest scores, performing on par with the references according to the CLIP score. The other open-source models fall far behind in terms of automatic scores. For the closed-source models, Midjourney performs best, closely followed by DALL-E-3. However, none of the models can match the TIFA accuracies of the reference images. Interestingly, even the human-created images don't achieve perfect accuracy. Yet, this could be due to the shortcomings of the models in the TIFA pipeline. 

The TIFA score is based on visual question answering, and the questions are categorized into the different elements that are evaluated. To dig deeper into the strengths and weaknesses of the T2I models, \autoref{fig:tifa_elements} shows the models' TIFA accuracies per element type. Most of the questions are targeted toward animals or humans, activities, and especially the objects depicted. The reference images (grey bars) outperform the image generation models, especially for animals/humans, activities, and attributes. This aligns with our assumption that body parts and movements are the hardest aspects for the models to generate. In contrast, almost all models outperform the reference images in terms of location and shape, and Stable Diffusion, as well as DALL-E, outperforms the references in the food category.

\subsection{Human evaluation}
\begin{table*}[ht]
    \centering
    \begin{tabular}{lcccc} \toprule
        \textbf{Model} & \textbf{Prompt coherence}$\uparrow$ & \textbf{Correctness}$\uparrow$ & \textbf{Bias}$\downarrow$ & \textbf{Suitability}$\uparrow$\\ \midrule
        \textbf{SD1\_4} & 0.48 ($\pm$ 0.80) & 0.19 ($\pm$ 0.45) & \textbf{0.00} ($\pm$ 0.00) & 0.06 ($\pm$ 0.29) \\
        \textbf{SD2\_1\_base} & 0.50 ($\pm$ 0.75) & 0.16 ($\pm$ 0.51) & \textbf{0.00} ($\pm$ 0.00) & 0.06 ($\pm$ 0.29) \\
        \textbf{SD\_3} & \textbf{1.48} ($\pm$ 0.98) & \textbf{0.90} ($\pm$ 0.94) & 0.14 ($\pm$ 0.61) & \textbf{0.51} ($\pm$ 0.83) \\
        \textbf{W\"urstchen} & 0.76 ($\pm$ 0.82) & 0.46 ($\pm$ 0.84) & \textbf{0.00} ($\pm$ 0.00) & 0.20 ($\pm$ 0.51) \\ 
        \midrule
        \textbf{DALL-E-3} & \textbf{2.23 }($\pm$ 0.91) & \textbf{2.19} ($\pm$ 0.96) & 0.21 ($\pm$ 0.74) & \textbf{1.85} ($\pm$ 1.12) \\
        \textbf{Midjourney} & 2.06 ($\pm$ 0.88)& 1.99 ($\pm$ 0.88) & 0.09 ($\pm$ 0.48) & 1.20 ($\pm$ 1.11) \\
        \textbf{Artbreeder} & 1.25 ($\pm$ 0.88)& 1.05 ($\pm$ 0.99) & \textbf{0.01} ($\pm$ 0.11) & 0.39 ($\pm$ 0.68) \\ 
        \bottomrule
    \end{tabular}
    \caption{Results from our human evaluation. The scores range from 0-3 and are averaged across all generated images. SD\_3 and DALL-E created the most accurate and most suitable images.}
    \label{tab:human_eval}
\end{table*}
While TIFA scores have a high correlation with human judgments \citep{hu-tifa}, automatic metrics can't cover all evaluation aspects. Especially for the overall correctness and simplicity of the images, there is currently no metric available.
Therefore, we added a human evaluation of our generated images. For this, we asked an expert for \ls{} (one of the authors) to manually review and rate the images. 
Images are an essential part of \ls{} \citep{din-spec-ls}, and many Easy language courses also address criteria for selecting appropriate images.
To reduce the overall workload, we selected one image per model and title, resulting in a dataset of 560 images. For each combination, we selected the image with the highest TIFA score. If two or more images shared the highest score, we sampled an image from among them.
The images were evaluated on four different scales by asking these questions:
\begin{itemize}
    \vspace{-0.1cm}\item \textit{Does the image follow the prompt?}: This question checks for missing or additional content. We only focused on relevant content and ignored aspects that did not affect the meaning of the image (e.g., the prompt describing a group of nine people, but the model only drew seven).
    \item \textit{Is the image correct?}: This question evaluates if the depicted content aligns with world knowledge, e.g., that people don't have three arms.
    \vspace{-0.15cm}\item \textit{Does the image exhibit a bias towards people with disabilities?}: This question is targeted towards the findings of previous work \citep{mack-bias-t2i,tevissen-disability-study} and evaluates whether the models tend to show people with disabilities as old or unhappy, even if the prompt does not define that.
    \vspace{-0.15cm}\item \textit{Is the image suitable for the target group?}: For the target group, it is important that the images are not overloaded with details, text, or colors and that they align with situations familiar to the target group. These criteria are in line with the DIN SPEC for \ls{} \citep{din-spec-ls}. In addition, this question checks whether the image is helpful to understand the original concept.
\end{itemize}
The human annotator could choose between four possible answers to the questions: no/indeterminable, partly, mostly, and yes. We mapped these answers to a numerical scale between 0 (answer no) and 3 (answer yes). The images were blinded, i.e., we only showed the annotator the images and the descriptions but not the name of the model that generated the image.

\autoref{tab:human_eval} shows the averaged scores from the human evaluation. While Stable Diffusion 3 outperformed the closed-source models in the automatic evaluation, it can not hold up to the expectation in the human evaluation, receiving significantly worse scores across all scales. Still, it is by far the best open-source model. The bad scores for the other open-source models are mostly due to unclear and indeterminable content. Remarkably, these open-source models show the least biases. However, this is an artifact from our evaluation setup: If the image does not show any depictable content, then it also can't show biases toward people with disabilities.

\begin{figure*}[ht]
    \centering
     \begin{subfigure}[b]{0.23\textwidth}
         \centering
         \includegraphics[width=\textwidth]{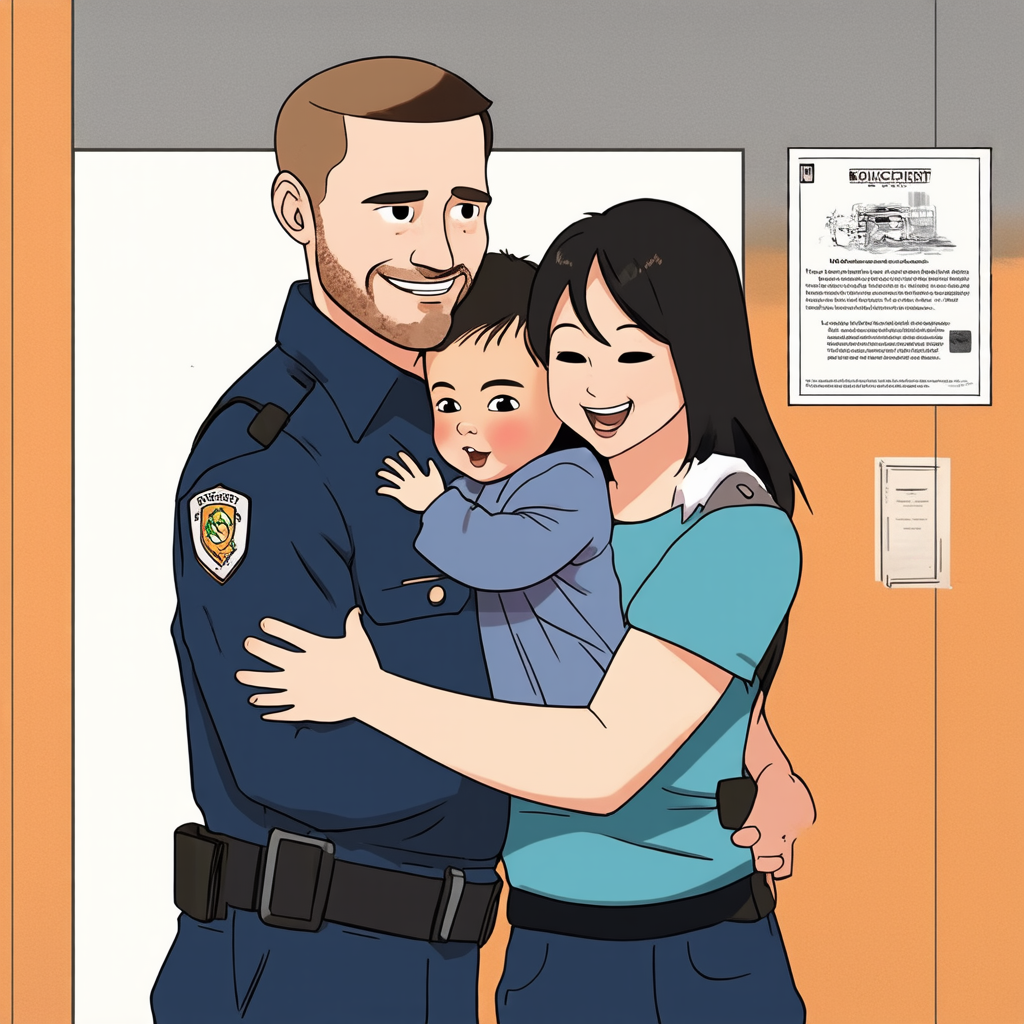}
         \caption{Cartoon picture of \textbf{Security} - Depicted are a woman and a man lovingly embracing a child.\\Created by SD3}
         \label{fig:example_sd3_security}
     \end{subfigure}
     \hfill
     \begin{subfigure}[b]{0.23\textwidth}
         \centering
         \includegraphics[width=\textwidth]{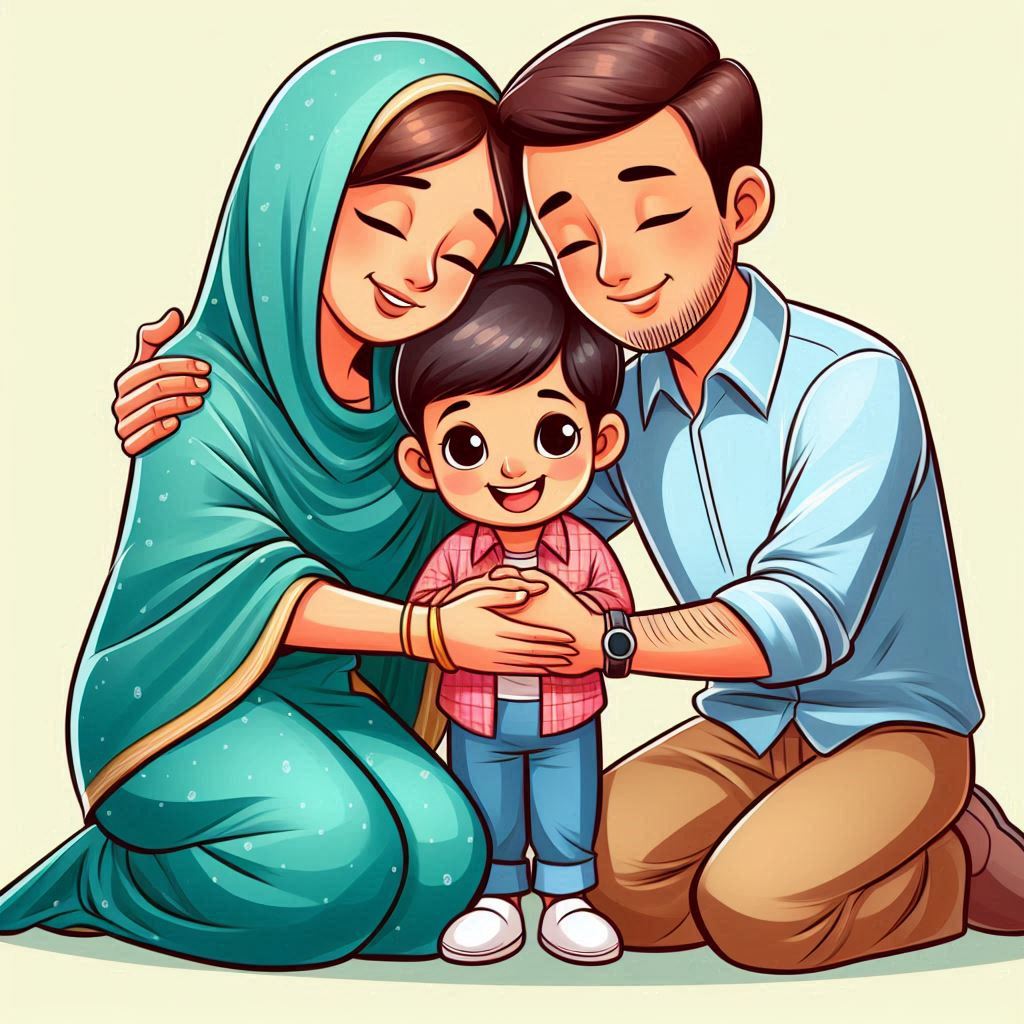}
         \caption{Cartoon picture of \textbf{Security} - Depicted are a woman and a man lovingly embracing a child.\\Created by DALL-E-3}
         \label{fig:example_dalle_security}
     \end{subfigure}
     \hfill
     \begin{subfigure}[b]{0.23\textwidth}
         \centering
         \includegraphics[width=\textwidth]{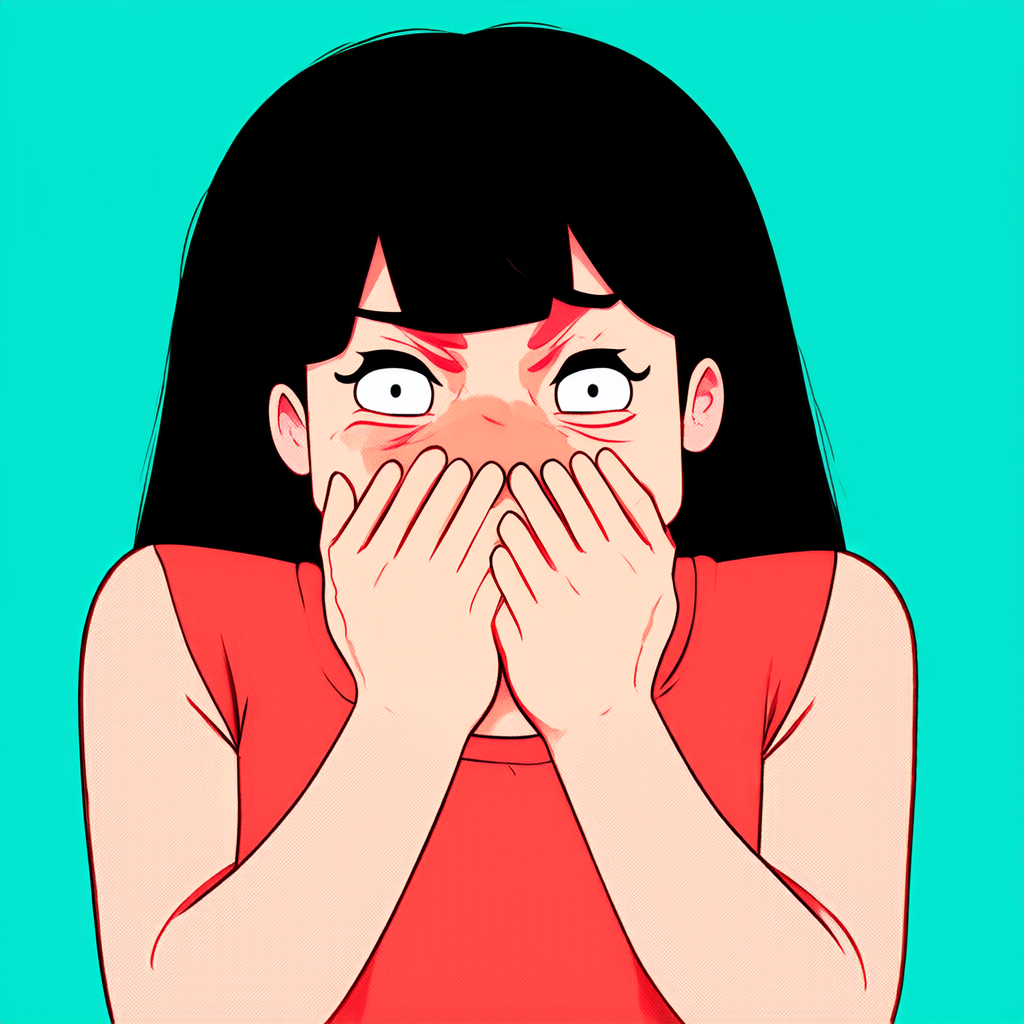}
         \caption{Cartoon picture of \textbf{Fear} - A woman covers her mouth with both hands. Her eyes and mouth are wide open. Sweat runs down her forehead.\\Created by SD3}
         \label{fig:example_sd3_fear}
     \end{subfigure}
     \hfill
     \begin{subfigure}[b]{0.23\textwidth}
         \centering
         \includegraphics[width=\textwidth]{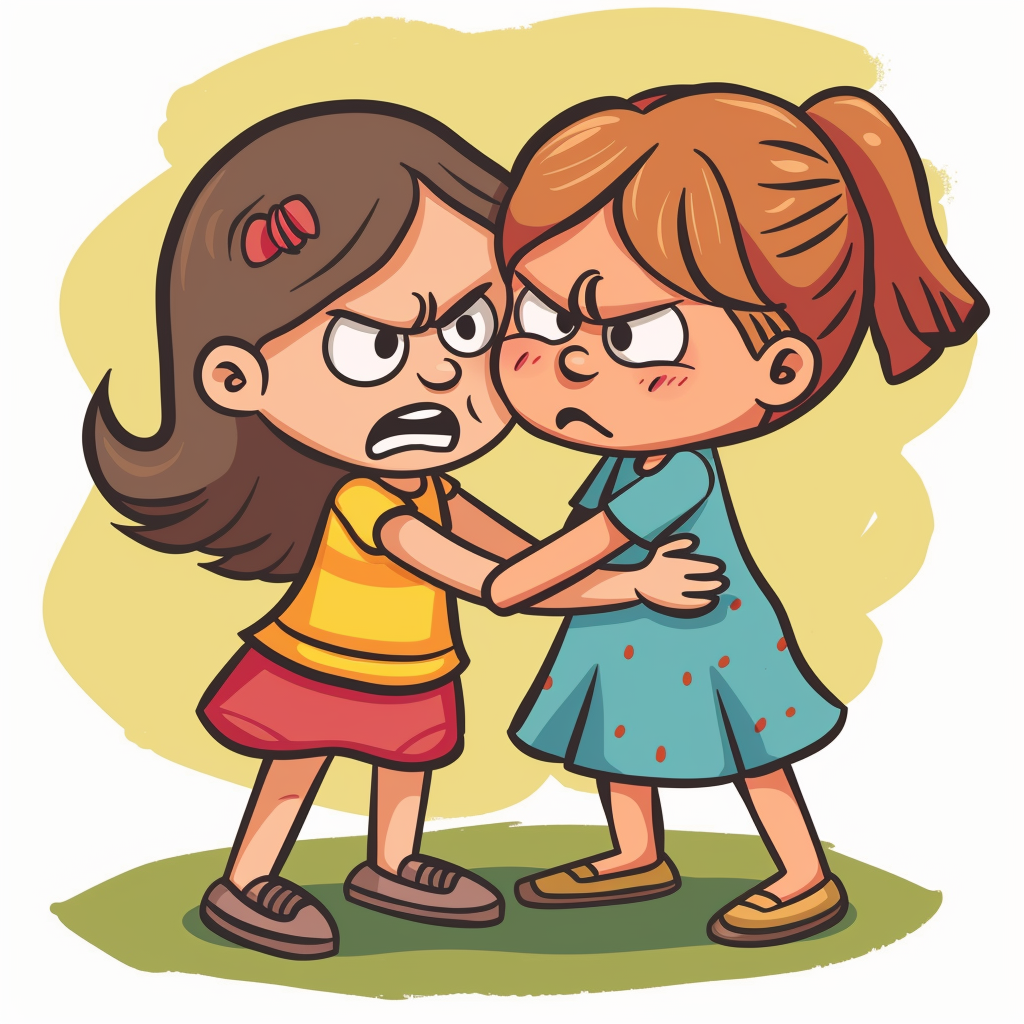}
         \caption{Cartoon picture of \textbf{Refusal of Physical Contact} - A woman tries to hug a girl. The girl looks away and resists.\\Created by Midjourney}
         \label{fig:example_midjourney_contact}
     \end{subfigure}
     \hfill
     \caption{Example prompts and generated images.}
     \label{fig:example_human_eval}
\end{figure*}

During our manual review, we made additional observations. Examples of them are depicted in \autoref{fig:example_human_eval}. The models sometimes hallucinate additional details. For example, one of the prompts is \enquote{Cartoon picture of Security - Depicted are a woman and a man lovingly embracing a child.}. Many models draw a policeman or officer, even though the prompt does not describe any (see \autoref{fig:example_sd3_security}). This indicates that the models have an inherent interpretation of world knowledge and, thus, associate security with police \citep{fu-Commonsense-T2I}. 

The biggest issues with the generated images arise with body parts and human motions. Examples are presented in Figures \ref{fig:example_sd3_security}, \ref{fig:example_sd3_fear} and \ref{fig:example_midjourney_contact} where body parts such as arms or legs are missing, or too many fingers were added. Another issue is that the models don't pay enough attention to small details, and thus, important aspects are missing. For example, for the prompt \enquote{Cartoon picture of Refusal of Eye Contact - A woman stands directly in front of a man and speaks to him. The man has his arms crossed in front of his chest. He does not look at her.}, all models created two people standing in front of one another, but no model could depict the refusal of eye contact properly. This could also be an issue with input token limits, i.e., that this important information was truncated. In addition, missing or misinterpreted details can change the meaning of the image. The images in Figures \ref{fig:example_sd3_fear} and \ref{fig:example_midjourney_contact} should show the emotions of fear and the refusal of physical contact. However, in both pictures, the people look rather angry and as if they would fight one another. Especially for people struggling with reading emotion from human expressions, this could evoke wrong associations. Therefore, such images are not suitable for the target group without further restrictions.

As indicated by the low bias scores in \autoref{tab:human_eval}, the models exhibit hardly any bias toward people with disabilities. The biases that we find are mostly related to hearing or vision impairments, where models tend to add an eye fold to visualize that a person is blind or draw incorrect hearing aids that look more like headsets. None of the models depicted people with disabilities as especially unhappy, except if the prompt especially stated it. On the contrary, most of them were smiling and happy.

The model with the best human evaluation scores is by far DALL-E-3. It was able to create correct images even for difficult body positions like in Yoga or hugging. In addition, the images were especially inclusive in terms of diversity: Pictures with multiple people often depicted people of color or people with glasses as parts of the groups. An example is \autoref{fig:example_dalle_security}, where the woman wears a head scarf, a garment only seen in minority groups in Western countries. These features were not described in the image prompts but added by the model and its world knowledge. 

\subsection{Feedback from the target group}
In line with the UN inclusion slogan \enquote{Nothing about us without us!} \citep{harpur-UN-convention} and in accordance with the DIN SPEC recommendation that the target group should review all content, we wanted to hear the opinion about the images from the target group. Therefore, we invited seven people with different disabilities (physical, mental, and combinations of both) between the ages of 21 and 42 for a workshop at the university. They were accompanied by their living assistants and two \ls{} experts. The study participants received a compensation of 32,50€ for their effort. We conducted two types of studies: comparative voting and a free-form discussion. Direct quotes from the participants are formatted in italic.

For the first part, we filtered the titles from the image dataset, where at least three models created suitable or mostly suitable images. Then, we selected seven titles from them. We presented all images with the same title at the same time but in a randomized order. The participants were asked to vote for all images they liked. We allowed multiple selections to account for equally good images. We used the voting platform Mentimeter\footnote{\url{https://www.mentimeter.com/}} where every participant can participate on their smartphones and submit their votes anonymously. With this, we could collect their opinions independently without being influenced by other participants. Some participants were supported by their living assistants when interacting with the smartphones.
The images and the participants' votes are presented in \autoref{tab:target_group_votes} in the Appendix.
\begin{figure*}[hbt]
    \centering
     \begin{subfigure}[b]{0.23\textwidth}
         \centering
         \includegraphics[width=\textwidth]{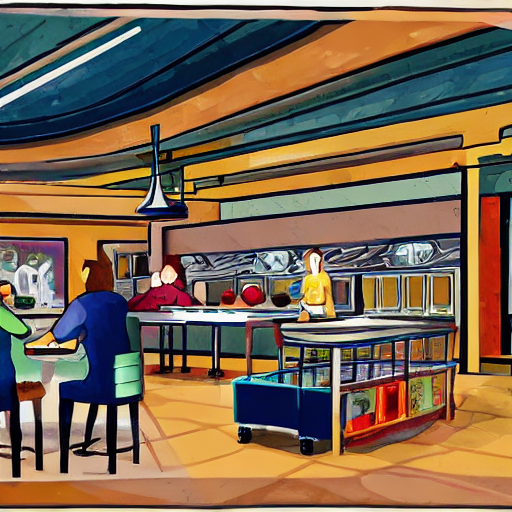}
         \caption{Cafeteria\\Created by SD1}
         \label{fig:example_sd1_cafeteria}
     \end{subfigure}
     \hfill
     \begin{subfigure}[b]{0.23\textwidth}
         \centering
         \includegraphics[width=\textwidth]{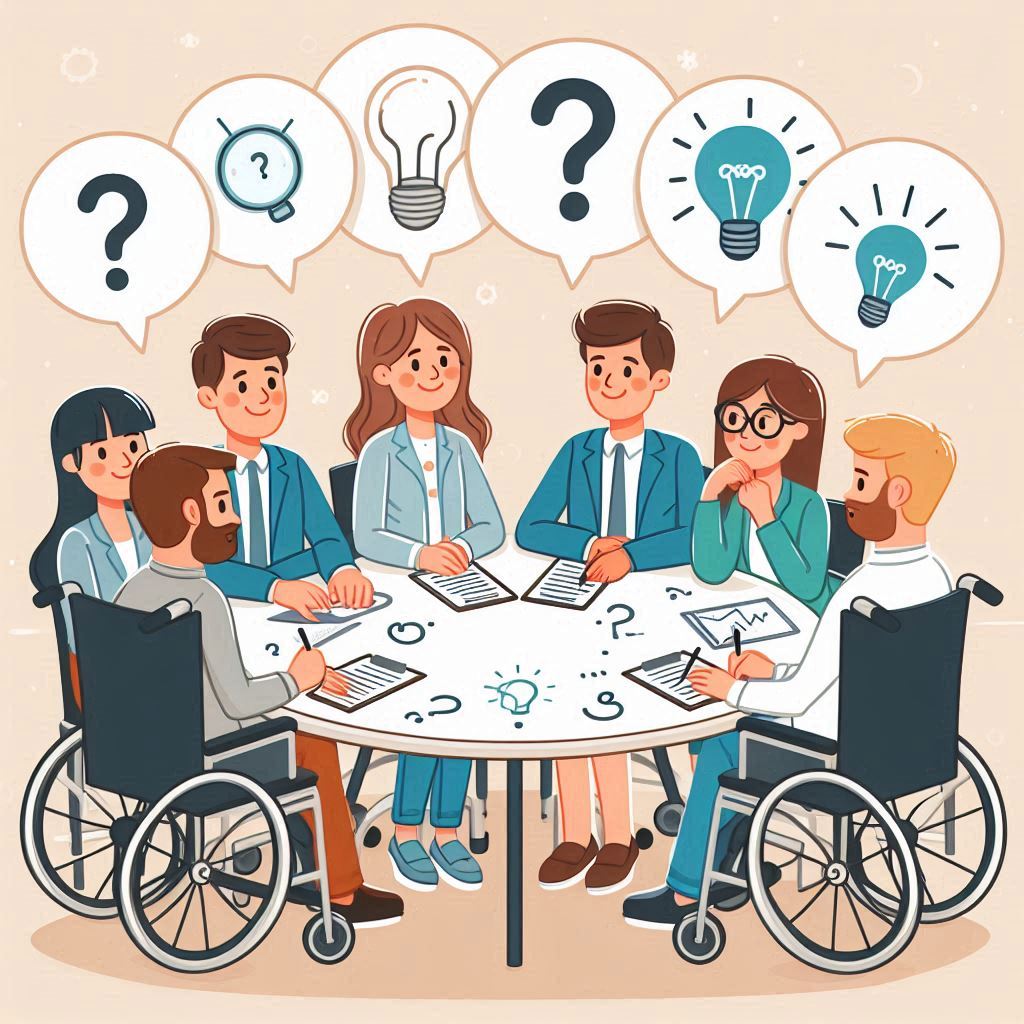}
         \caption{Brainstorming\\Created by DALL-E-3}
         \label{fig:example_dalle_brainstorming}
     \end{subfigure}
     \hfill
     \begin{subfigure}[b]{0.23\textwidth}
         \centering
         \includegraphics[width=\textwidth]{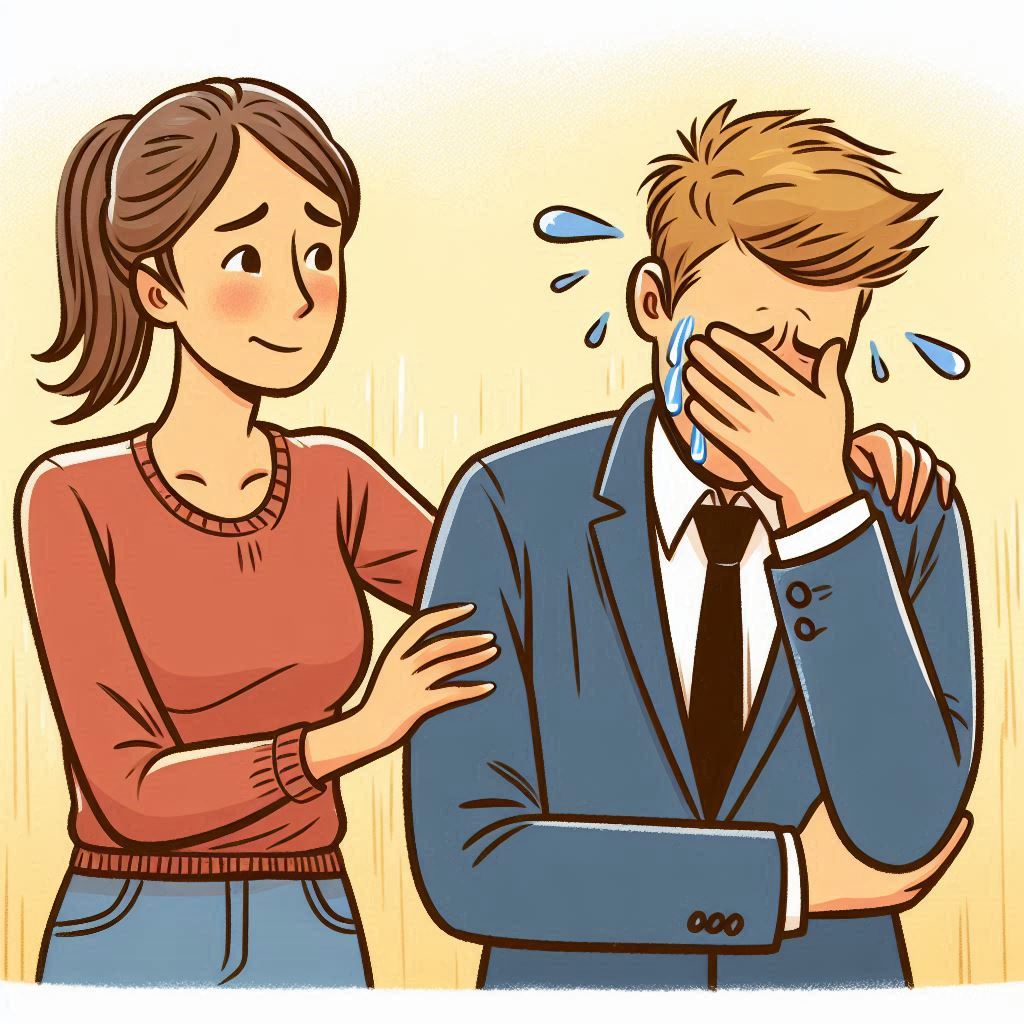}
         \caption{Empathy\\Created by DALL-E-3}
         \label{fig:example_dalle_empathy}
     \end{subfigure}
     \hfill
     \begin{subfigure}[b]{0.23\textwidth}
         \centering
         \includegraphics[width=\textwidth]{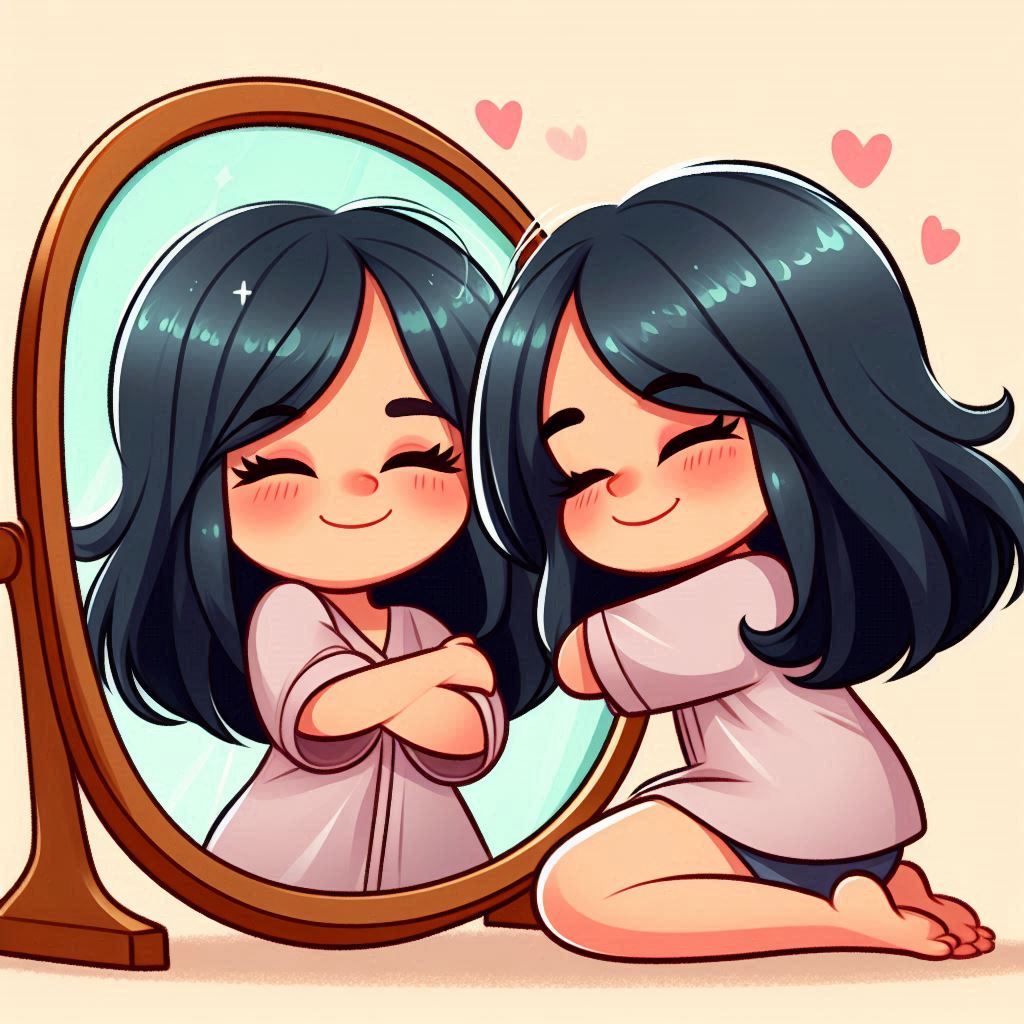}
         \caption{Self-love\\Created by DALL-E-3}
         \label{fig:example_dalle_selflove}
     \end{subfigure}
     \hfill
     \caption{Term-guessing images. The images were shown to the target group participants, and they had to guess the title or describe what the word could mean if they didn't know it.}
     \label{fig:term_guessing_target}
\end{figure*}

For most of the images, DALL-E received the most votes, often even more than the reference images. In general, colorful images with few additional or decorative content were preferred. Some images seemed a bit abstract, e.g., the fruit depicted for the bowl of vitamins had a few mistakes or weird coloring. Nevertheless, the participants showed great creativity when naming the fruits. Therefore, if the context is clear, small mistakes don't bother too much. This also came clear when participants explained why they chose certain images for the bedroom: they chose the ones that looked the most \enquote{\textit{cozy}}. In contrast, some toothbrushes in the hygiene product images looked unrealistic, not suitable for the teeth, or were simply wrong. Multiple participants were distracted by these mistakes and started a discussion about what was wrong with the toothbrushes and how it would hurt to actually use them.

The second part of the workshop was more open to direct feedback. We presented four different images, one from Stable Diffusion 1 and three from DALL-E-3, and the participants should guess the depicted content. The images and their titles are shown in \autoref{fig:term_guessing_target}. We chose some complex terms on purpose to assess whether the images could help with understanding them. For example, the word \enquote{Brainstorming} was unfamiliar to half of the participants. Nevertheless, they could describe and explain the word as \enquote{\textit{people are sitting together and collect ideas}} only based on the image. This shows that the selected images are not only suitable for illustrating texts, but they also fulfill their purpose of explaining complex terms with ease.

In addition to the term guessing of the second part, the participants were also invited to express their thoughts and opinions about the presented images. We try to summarize them in the following:
\begin{itemize}
    \vspace{-0.15cm}\item Participants did not like black\&white-only images because \enquote{\textit{it makes you depressed}}.
    \vspace{-0.15cm}\item Different illustrations of the same objects (e.g., the light bulbs in the brainstorming image) were confusing, and participants tried to find a reason for the differences, even if there was no reason for that.
    \vspace{-0.15cm}\item The images should show accessible situations, i.e., suitable for people using wheelchairs or hearing aids. We had a discussion about whether the counter in SD1's cafeteria was accessible for wheelchairs and whether people would need help to reach all the offers. For this discussion, the blurry and abstract style of the image was of minor relevance.
    \vspace{-0.15cm}\item In the DALL-E-3 image for yoga (\autoref{tab:target_group_votes} in Appendix), the person using the wheelchair is very old. When the two participants who used a wheelchair were asked whether they felt discriminated by this, they answered \enquote{\textit{No, why should I? Even old people can do yoga!}}
\end{itemize}


\section{Conclusion}
In this paper, we have explored whether text-to-image models can be utilized to create illustrations for easy-to-read texts. For this, we evaluated the generated images in large-scale human studies, including seven participants from the target group. Closed-source models like DALL-E-3 and Midjourney and the open-source Stable Diffusion 3 have shown impressive performance in creating these images, sometimes creating even more favorable images than the gold-standard references. However, their performance highly depends on the depicted content, and the models struggle with difficult postures and specific body parts especially. Therefore, they cannot be used without human oversight or multiple iterations of image description optimization. In addition, the best-performing models are closed-source or very large in parameter size, meaning that text creators will still have to pay for their images. Since text creators will use the models, they have human expert oversight, and thus, every generated image will be reviewed, and erroneous images can be filtered before being shown to the target group. Finally, we believe that T2I models are especially suited for accessible communication due to their fast availability and options for tailored, customizable, and copyright-free content. 

In future research, we would like to get rid of the intermediate step of explicit image descriptions and hope to see models that can create the images directly from the text paragraph. In addition, we would like to investigate their compliance with prompts in German and other non-English languages and investigate conditioning the images on the reference images during generation.

\section*{Limitations and ethical considerations}
Our work presents a quite extensive comparison of different T2I images. While we did our best to include as many models with different architectures, sizes, and availabilities, we can only test the models published by the time of writing this paper. The current developments and improvements in AI are rapid, and thus, there may be newer and better models soon that we couldn't include in our study.

We tried to design this study as participatory as possible and included seven people from the target group in our human evaluation. They received 32,50€ to compensate for their effort. Nevertheless, the feedback session was moderated, and the authors pre-selected the images. A target group evaluation of all images would be infeasible and not of any help to the target group. Still, our image selection and moderation introduced a bias from the authors on them that we can not neglect. In addition, the disabilities and needs of the target group are very diverse and cannot be represented by only seven people. Nevertheless, we try to make their opinions be heard and invite all researchers in the area of accessible communication to work together with the target group.

Finally, we are aware that generative AI, whether it generates text, images, or any other modality, is being criticized for threatening jobs and content quality. The goal of our work is in no way to replace humans in the process of creating accessible content. However, we believe that the benefits of the short-time availability of simplified texts and images are important to overcome information barriers, especially on the internet. Studies such as ours can be of great help to further improve the quality of those models and to align their objectives with what is actually needed by the target group. In the end, our investigations show that the T2I models are far from being perfect and still need careful human oversight. Especially in terms of image evaluation, we could not find an automatic metric that was satisfactory in alignment with our judgment.

\section*{Lay Summary}
Creating texts that are easy to read and understand is important for people with disabilities, learning difficulties, or those who have trouble with reading. These easy-to-read (E2R) texts often include pictures to help explain the information. However, it can be hard to find images that fit the specific needs of each text. Hiring artists to make customized images can be expensive, and existing image databases don’t allow for easy changes to match the content of the text.

Our study looks at whether we can use artificial intelligence (AI) to generate these images quickly and cheaply. We tested seven different AI tools, called text-to-image models, which create pictures based on written descriptions. Some of these tools are open to the public, while others are not. We wanted to see if these AI-generated images could be a good solution for E2R creators.

We evaluated over 2,000 images created by these models and manually reviewed 560 of them. During the review, we looked at how well the images matched the description, if they were accurate, if they had any bias against people with disabilities, and if they were useful for the target group. Our results show that while some models produced high-quality images, none of them are ready to be used on a large scale without human oversight.

We also conducted a user study with seven people from the E2R target group to gather feedback on how well the images met their needs. It is important to include the target group and their opinions and preferences when doing research. The feedback was helpful in identifying areas where the AI models worked well and where they fell short.

Our research is an important first step toward making it easier and more affordable to create images that help make information more accessible. However, more improvements are needed before these AI tools can fully replace human involvement in creating custom images for E2R texts.

\section*{Acknowledgments}
The images were provided by the Landesarbeitsgemeinschaft Selbsthilfe von Menschen mit Behinderungen und chronischen Erkrankungen Rheinland-Pfalz e.V. with the kind support of AOK Rheinland-Pfalz/Saarland. We thank them for their assistance that made our research possible in the first place!\\
In addition, we thank the \textit{FÜP - FortSchritt Übersetzungs- \& Prüfbüro für Leichte Sprache} (office for \ls{} translation and evaluation) by the FortSchritt Verein zur Verbreitung der Konduktiven Förderung e.V.\footnote{\url{https://www.fortschritt-bayern.de/angebote/leichte-sprache}} and especially their review group for the insightful feedback.\\
This research has been funded by the German Federal Ministry of Education and Research (BMBF) through grant 01IS23069 Software Campus 3.0 (Technical University of Munich) as part of the Software Campus project \enquote{LIANA}.\\
This paper is based on a joined work in the context of Tringa Sylaj's master’s thesis \cite{Sylaj-MA-image-gen}.

\bibliography{custom}

\begin{thebibliography}{34}
\providecommand{\natexlab}[1]{#1}

\bibitem[{Ansch{\"u}tz et~al.(2023)Ansch{\"u}tz, Oehms, Wimmer, Jezierski, and
  Groh}]{anschutz-german-simp}
Miriam Ansch{\"u}tz, Joshua Oehms, Thomas Wimmer, Bart{\l}omiej Jezierski, and
  Georg Groh. 2023.
\newblock \href {https://doi.org/10.18653/v1/2023.findings-acl.74} {Language
  models for {G}erman text simplification: Overcoming parallel data scarcity
  through style-specific pre-training}.
\newblock In \emph{Findings of the Association for Computational Linguistics:
  ACL 2023}, pages 1147--1158, Toronto, Canada. Association for Computational
  Linguistics.

\bibitem[{Barratt and Sharma(2018)}]{barratt-inception-critique}
Shane Barratt and Rishi Sharma. 2018.
\newblock \href {https://arxiv.org/abs/1801.01973} {A note on the inception
  score}.
\newblock \emph{Preprint}, arXiv:1801.01973.

\bibitem[{Brock et~al.(2019)Brock, Donahue, and
  Simonyan}]{brock2019largescalegantraining}
Andrew Brock, Jeff Donahue, and Karen Simonyan. 2019.
\newblock \href {https://openreview.net/forum?id=B1xsqj09Fm} {Large scale {GAN}
  training for high fidelity natural image synthesis}.
\newblock In \emph{International Conference on Learning Representations}.

\bibitem[{Das et~al.(2024)Das, Fiannaca, Morris, Kane, and
  Bennett}]{das-alt-text-study}
Maitraye Das, Alexander~J. Fiannaca, Meredith~Ringel Morris, Shaun~K. Kane, and
  Cynthia~L. Bennett. 2024.
\newblock \href {https://doi.org/10.1145/3613904.3642325} {From provenance to
  aberrations: Image creator and screen reader user perspectives on alt text
  for ai-generated images}.
\newblock In \emph{Proceedings of the CHI Conference on Human Factors in
  Computing Systems}, CHI '24, New York, NY, USA. Association for Computing
  Machinery.

\bibitem[{{DIN-Normenausschuss Ergonomie}(2023)}]{din-spec-ls}
{DIN-Normenausschuss Ergonomie}. 2023.
\newblock \href {https://doi.org/https://dx.doi.org/10.31030/3417293}
  {Empfehlungen für {Deutsche} {Leichte} {Sprache} ({DIN} {SPEC} 33429)}.

\bibitem[{Fu et~al.(2024)Fu, He, Lu, Wang, and Roth}]{fu-Commonsense-T2I}
Xingyu Fu, Muyu He, Yujie Lu, William~Yang Wang, and Dan Roth. 2024.
\newblock \href {https://openreview.net/forum?id=MI52iXSSNy} {Commonsense-t2i
  challenge: Can text-to-image generation models understand commonsense?}
\newblock In \emph{First Conference on Language Modeling}.

\bibitem[{Geislinger et~al.(2023)Geislinger, Pourasad, G{\"u}l, Djahangir,
  Muhie~Yimam, Remus, and Biemann}]{geislinger-l2learners-eyetracking}
Robert Geislinger, Ali~Ebrahimi Pourasad, Deniz G{\"u}l, Daniel Djahangir, Seid
  Muhie~Yimam, Steffen Remus, and Chris Biemann. 2023.
\newblock \href {https://aclanthology.org/2023.limo-1.2} {Multi-modal learning
  application {--} support language learners with {NLP} techniques and
  eye-tracking}.
\newblock In \emph{Proceedings of the 1st Workshop on Linguistic Insights from
  and for Multimodal Language Processing}, pages 6--11, Ingolstadt, Germany.
  Association for Computational Lingustics.

\bibitem[{Goodfellow et~al.(2020)Goodfellow, Pouget-Abadie, Mirza, Xu,
  Warde-Farley, Ozair, Courville, and
  Bengio}]{goodfellow2014generativeadversarialnetworks}
Ian Goodfellow, Jean Pouget-Abadie, Mehdi Mirza, Bing Xu, David Warde-Farley,
  Sherjil Ozair, Aaron Courville, and Yoshua Bengio. 2020.
\newblock \href {https://doi.org/10.1145/3422622} {Generative adversarial
  networks}.
\newblock \emph{Commun. ACM}, 63(11):139–144.

\bibitem[{Harpur and Stein(2017)}]{harpur-UN-convention}
Paul Harpur and Michael~Ashley Stein. 2017.
\newblock The convention on the rights of persons with disabilities as a global
  tipping point for the participation of persons with disabilities.
\newblock In \emph{Oxford Research Encyclopedia of Politics}.

\bibitem[{Hessel et~al.(2021)Hessel, Holtzman, Forbes, Le~Bras, and
  Choi}]{hessel-clip}
Jack Hessel, Ari Holtzman, Maxwell Forbes, Ronan Le~Bras, and Yejin Choi. 2021.
\newblock \href {https://doi.org/10.18653/v1/2021.emnlp-main.595} {{CLIPS}core:
  A reference-free evaluation metric for image captioning}.
\newblock In \emph{Proceedings of the 2021 Conference on Empirical Methods in
  Natural Language Processing}, pages 7514--7528, Online and Punta Cana,
  Dominican Republic. Association for Computational Linguistics.

\bibitem[{Hu et~al.(2023)Hu, Liu, Kasai, Wang, Ostendorf, Krishna, and
  Smith}]{hu-tifa}
Y.~Hu, B.~Liu, J.~Kasai, Y.~Wang, M.~Ostendorf, R.~Krishna, and N.~A. Smith.
  2023.
\newblock \href {https://doi.org/10.1109/ICCV51070.2023.01866} {Tifa: Accurate
  and interpretable text-to-image faithfulness evaluation with question
  answering}.
\newblock In \emph{2023 IEEE/CVF International Conference on Computer Vision
  (ICCV)}, pages 20349--20360, Los Alamitos, CA, USA. IEEE Computer Society.

\bibitem[{Huh et~al.(2023)Huh, Peng, and Pavel}]{huh-GenAssist}
Mina Huh, Yi-Hao Peng, and Amy Pavel. 2023.
\newblock \href {https://doi.org/10.1145/3586183.3606735} {Genassist: Making
  image generation accessible}.
\newblock In \emph{Proceedings of the 36th Annual ACM Symposium on User
  Interface Software and Technology}, UIST '23, New York, NY, USA. Association
  for Computing Machinery.

\bibitem[{James~Edwards et~al.(2021)James~Edwards, Lewis~Polster, Tuason,
  Blank, Gilbert, and Branham}]{edwards-study-image-description}
Emory James~Edwards, Kyle Lewis~Polster, Isabel Tuason, Emily Blank, Michael
  Gilbert, and Stacy Branham. 2021.
\newblock \href {https://doi.org/10.1145/3441852.3471222} {"that's in the eye
  of the beholder": Layers of interpretation in image descriptions for
  fictional representations of people with disabilities}.
\newblock In \emph{Proceedings of the 23rd International ACM SIGACCESS
  Conference on Computers and Accessibility}, ASSETS '21, New York, NY, USA.
  Association for Computing Machinery.

\bibitem[{Karras et~al.(2021)Karras, Laine, and
  Aila}]{karras2019stylebasedgeneratorarchitecturegenerative}
Tero Karras, Samuli Laine, and Timo Aila. 2021.
\newblock \href {https://doi.org/10.1109/TPAMI.2020.2970919} {A style-based
  generator architecture for generative adversarial networks}.
\newblock \emph{IEEE Trans. Pattern Anal. Mach. Intell.}, 43(12):4217–4228.

\bibitem[{Khashabi et~al.(2020)Khashabi, Min, Khot, Sabharwal, Tafjord, Clark,
  and Hajishirzi}]{khashabi-unifiedqa}
Daniel Khashabi, Sewon Min, Tushar Khot, Ashish Sabharwal, Oyvind Tafjord,
  Peter Clark, and Hannaneh Hajishirzi. 2020.
\newblock \href {https://doi.org/10.18653/v1/2020.findings-emnlp.171}
  {{UNIFIEDQA}: Crossing format boundaries with a single {QA} system}.
\newblock In \emph{Findings of the Association for Computational Linguistics:
  EMNLP 2020}, pages 1896--1907, Online. Association for Computational
  Linguistics.

\bibitem[{Kiesel et~al.(2024)Kiesel, {\c C}{\"o}ltekin, Heinrich, Fr{\"o}be,
  Alshomary, Longueville, Erjavec, Handke, Kopp, Ljube\v{s}i\'{c}, Meden,
  Mirzakhmedova, Morkevi\v{c}ius, Reitis-M{\"u}nstermann, Scharfbillig,
  Stefanovitch, Wachsmuth, Potthast, and Stein}]{kiesel-clef-touche}
Johannes Kiesel, {\c C}a{\u g}r{\i} {\c C}{\"o}ltekin, Maximilian Heinrich,
  Maik Fr{\"o}be, Milad Alshomary, Bertrand~De Longueville, Toma\v{z} Erjavec,
  Nicolas Handke, Maty\'{a}\v{s} Kopp, Nikola Ljube\v{s}i\'{c}, Katja Meden,
  Nailia Mirzakhmedova, Vaidas Morkevi\v{c}ius, Theresa Reitis-M{\"u}nstermann,
  Mario Scharfbillig, Nicolas Stefanovitch, Henning Wachsmuth, Martin Potthast,
  and Benno Stein. 2024.
\newblock \href {https://doi.org/10.1007/978-3-031-71908-0_14} {{Overview of
  Touch{\'e} 2024: Argumentation Systems}}.
\newblock In \emph{Experimental IR Meets Multilinguality, Multimodality, and
  Interaction. 15th International Conference of the CLEF Association (CLEF
  2024)}, volume 14959 of \emph{Lecture Notes in Computer Science}, pages
  308--332, Berlin Heidelberg New York. Springer.

\bibitem[{Li et~al.(2022)Li, Xu, Tian, Wang, Yan, Bi, Ye, Chen, Xu, Cao, Zhang,
  Huang, Huang, Zhou, and Si}]{li-mplug}
Chenliang Li, Haiyang Xu, Junfeng Tian, Wei Wang, Ming Yan, Bin Bi, Jiabo Ye,
  He~Chen, Guohai Xu, Zheng Cao, Ji~Zhang, Songfang Huang, Fei Huang, Jingren
  Zhou, and Luo Si. 2022.
\newblock \href {https://doi.org/10.18653/v1/2022.emnlp-main.488} {m{PLUG}:
  Effective and efficient vision-language learning by cross-modal
  skip-connections}.
\newblock In \emph{Proceedings of the 2022 Conference on Empirical Methods in
  Natural Language Processing}, pages 7241--7259, Abu Dhabi, United Arab
  Emirates. Association for Computational Linguistics.

\bibitem[{Mack et~al.(2024)Mack, Qadri, Denton, Kane, and
  Bennett}]{mack-bias-t2i}
Kelly~Avery Mack, Rida Qadri, Remi Denton, Shaun~K. Kane, and Cynthia~L.
  Bennett. 2024.
\newblock \href {https://doi.org/10.1145/3613904.3642166} {“they only care to
  show us the wheelchair”: disability representation in text-to-image ai
  models}.
\newblock In \emph{Proceedings of the CHI Conference on Human Factors in
  Computing Systems}, CHI '24, New York, NY, USA. Association for Computing
  Machinery.

\bibitem[{Madina et~al.(2023)Madina, Gonzalez-Dios, and Siegel}]{madina-e2r}
Margot Madina, Itziar Gonzalez-Dios, and Melanie Siegel. 2023.
\newblock \href {https://doi.org/10.1145/3594806.3596530} {Easy-to-read
  language resources and tools for three european languages}.
\newblock In \emph{Proceedings of the 16th International Conference on
  PErvasive Technologies Related to Assistive Environments}, PETRA '23, page
  693–699, New York, NY, USA. Association for Computing Machinery.

\bibitem[{OpenAI et~al.(2024)OpenAI, Achiam, Adler, Agarwal, Ahmad, Akkaya,
  Aleman, Almeida, Altenschmidt, Altman, Anadkat, Avila, Babuschkin, Balaji,
  Balcom, Baltescu, Bao, Bavarian, Belgum, Bello, Berdine, Bernadett-Shapiro,
  Berner, Bogdonoff, and et~al.}]{openai-gpt4}
OpenAI, Josh Achiam, Steven Adler, Sandhini Agarwal, Lama Ahmad, Ilge Akkaya,
  Florencia~Leoni Aleman, Diogo Almeida, Janko Altenschmidt, Sam Altman,
  Shyamal Anadkat, Red Avila, Igor Babuschkin, Suchir Balaji, Valerie Balcom,
  Paul Baltescu, Haiming Bao, Mohammad Bavarian, Jeff Belgum, Irwan Bello, Jake
  Berdine, Gabriel Bernadett-Shapiro, Christopher Berner, Lenny Bogdonoff, and
  et~al. 2024.
\newblock \href {https://arxiv.org/abs/2303.08774} {Gpt-4 technical report}.
\newblock \emph{Preprint}, arXiv:2303.08774.

\bibitem[{Pernias et~al.(2024)Pernias, Rampas, Richter, Pal, and
  Aubreville}]{pernias2023wuerstchen}
Pablo Pernias, Dominic Rampas, Mats~Leon Richter, Christopher Pal, and Marc
  Aubreville. 2024.
\newblock \href {https://openreview.net/forum?id=gU58d5QeGv} {W\"urstchen: An
  efficient architecture for large-scale text-to-image diffusion models}.
\newblock In \emph{The Twelfth International Conference on Learning
  Representations}.

\bibitem[{Proven-Bessel et~al.(2021)Proven-Bessel, Zhao, and
  Chen}]{provenbessel-ComicGAN}
Ben Proven-Bessel, Zilong Zhao, and Lydia Chen. 2021.
\newblock \href {https://arxiv.org/abs/2109.09120} {Comicgan: Text-to-comic
  generative adversarial network}.
\newblock \emph{Preprint}, arXiv:2109.09120.

\bibitem[{Radford et~al.(2021)Radford, Kim, Hallacy, Ramesh, Goh, Agarwal,
  Sastry, Askell, Mishkin, Clark, Krueger, and Sutskever}]{radford-CLIP}
Alec Radford, Jong~Wook Kim, Chris Hallacy, Aditya Ramesh, Gabriel Goh,
  Sandhini Agarwal, Girish Sastry, Amanda Askell, Pamela Mishkin, Jack Clark,
  Gretchen Krueger, and Ilya Sutskever. 2021.
\newblock \href {https://proceedings.mlr.press/v139/radford21a.html} {Learning
  transferable visual models from natural language supervision}.
\newblock In \emph{Proceedings of the 38th International Conference on Machine
  Learning}, volume 139 of \emph{Proceedings of Machine Learning Research},
  pages 8748--8763. PMLR.

\bibitem[{Ramesh et~al.(2021)Ramesh, Pavlov, Goh, Gray, Voss, Radford, Chen,
  and Sutskever}]{ramesh2021zeroshot}
Aditya Ramesh, Mikhail Pavlov, Gabriel Goh, Scott Gray, Chelsea Voss, Alec
  Radford, Mark Chen, and Ilya Sutskever. 2021.
\newblock \href {https://proceedings.mlr.press/v139/ramesh21a.html} {Zero-shot
  text-to-image generation}.
\newblock In \emph{Proceedings of the 38th International Conference on Machine
  Learning}, volume 139 of \emph{Proceedings of Machine Learning Research},
  pages 8821--8831. PMLR.

\bibitem[{Rombach et~al.(2022)Rombach, Blattmann, Lorenz, Esser, and
  Ommer}]{rombach2022highresolution}
Robin Rombach, Andreas Blattmann, Dominik Lorenz, Patrick Esser, and Bj\"orn
  Ommer. 2022.
\newblock High-resolution image synthesis with latent diffusion models.
\newblock In \emph{Proceedings of the IEEE/CVF Conference on Computer Vision
  and Pattern Recognition (CVPR)}, pages 10684--10695.

\bibitem[{Salimans et~al.(2016)Salimans, Goodfellow, Zaremba, Cheung, Radford,
  Chen, and Chen}]{Salimans-Inception-score}
Tim Salimans, Ian Goodfellow, Wojciech Zaremba, Vicki Cheung, Alec Radford,
  Xi~Chen, and Xi~Chen. 2016.
\newblock \href
  {https://proceedings.neurips.cc/paper_files/paper/2016/file/8a3363abe792db2d8761d6403605aeb7-Paper.pdf}
  {Improved techniques for training gans}.
\newblock In \emph{Advances in Neural Information Processing Systems},
  volume~29. Curran Associates, Inc.

\bibitem[{Schneider et~al.(2021)Schneider, Ala{\c{c}}am, Wang, and
  Biemann}]{schneider-retrieval-textbook-benchmark}
Florian Schneider, {\"O}zge Ala{\c{c}}am, Xintong Wang, and Chris Biemann.
  2021.
\newblock \href {https://aclanthology.org/2021.naacl-srw.21} {Towards
  multi-modal text-image retrieval to improve human reading}.
\newblock In \emph{Proceedings of the 2021 Conference of the North American
  Chapter of the Association for Computational Linguistics: Student Research
  Workshop}, Online. Association for Computational Linguistics.

\bibitem[{Schomacker et~al.(2023)Schomacker, Gille, Tropmann-Frick, and von~der
  H{\"u}lls}]{schomacker-accessible}
Thorben Schomacker, Michael Gille, Marina Tropmann-Frick, and J{\"o}rg von~der
  H{\"u}lls. 2023.
\newblock \href {https://aclanthology.org/2023.konvens-main.6} {Data and
  approaches for {G}erman text simplification {--} towards an
  accessibility-enhanced communication}.
\newblock In \emph{Proceedings of the 19th Conference on Natural Language
  Processing (KONVENS 2023)}, pages 63--68, Ingolstadt, Germany. Association
  for Computational Lingustics.

\bibitem[{Singh et~al.(2023)Singh, Zouhar, and Sachan}]{singh-textbook-images}
Janvijay Singh, Vil{\'e}m Zouhar, and Mrinmaya Sachan. 2023.
\newblock \href {https://doi.org/10.18653/v1/2023.emnlp-main.731} {Enhancing
  textbooks with visuals from the web for improved learning}.
\newblock In \emph{Proceedings of the 2023 Conference on Empirical Methods in
  Natural Language Processing}, pages 11931--11944, Singapore. Association for
  Computational Linguistics.

\bibitem[{Stodden et~al.(2023)Stodden, Momen, and Kallmeyer}]{stodden-deplain}
Regina Stodden, Omar Momen, and Laura Kallmeyer. 2023.
\newblock \href {https://doi.org/10.18653/v1/2023.acl-long.908} {{DE}plain: A
  {G}erman parallel corpus with intralingual translations into plain language
  for sentence and document simplification}.
\newblock In \emph{Proceedings of the 61st Annual Meeting of the Association
  for Computational Linguistics (Volume 1: Long Papers)}, pages 16441--16463,
  Toronto, Canada. Association for Computational Linguistics.

\bibitem[{Tevissen(2024)}]{tevissen-disability-study}
Yannis Tevissen. 2024.
\newblock \href {https://arxiv.org/abs/2406.14993} {Disability representations:
  Finding biases in automatic image generation}.
\newblock \emph{Preprint}, arXiv:2406.14993.

\bibitem[{Touvron et~al.(2023)Touvron, Martin, Stone, Albert, Almahairi,
  Babaei, Bashlykov, Batra, Bhargava, Bhosale, Bikel, Blecher, Ferrer, Chen,
  Cucurull, Esiobu, Fernandes, Fu, Fu, Fuller, Gao, Goswami, Goyal, Hartshorn,
  Hosseini, Hou, Inan, Kardas, Kerkez, Khabsa, Kloumann, Korenev, Koura,
  Lachaux, Lavril, Lee, Liskovich, Lu, Mao, Martinet, Mihaylov, Mishra,
  Molybog, Nie, Poulton, Reizenstein, Rungta, Saladi, Schelten, Silva, Smith,
  Subramanian, Tan, Tang, Taylor, Williams, Kuan, Xu, Yan, Zarov, Zhang, Fan,
  Kambadur, Narang, Rodriguez, Stojnic, Edunov, and Scialom}]{touvron-llama2}
Hugo Touvron, Louis Martin, Kevin Stone, Peter Albert, Amjad Almahairi, Yasmine
  Babaei, Nikolay Bashlykov, Soumya Batra, Prajjwal Bhargava, Shruti Bhosale,
  Dan Bikel, Lukas Blecher, Cristian~Canton Ferrer, Moya Chen, Guillem
  Cucurull, David Esiobu, Jude Fernandes, Jeremy Fu, Wenyin Fu, Brian Fuller,
  Cynthia Gao, Vedanuj Goswami, Naman Goyal, Anthony Hartshorn, Saghar
  Hosseini, Rui Hou, Hakan Inan, Marcin Kardas, Viktor Kerkez, Madian Khabsa,
  Isabel Kloumann, Artem Korenev, Punit~Singh Koura, Marie-Anne Lachaux,
  Thibaut Lavril, Jenya Lee, Diana Liskovich, Yinghai Lu, Yuning Mao, Xavier
  Martinet, Todor Mihaylov, Pushkar Mishra, Igor Molybog, Yixin Nie, Andrew
  Poulton, Jeremy Reizenstein, Rashi Rungta, Kalyan Saladi, Alan Schelten, Ruan
  Silva, Eric~Michael Smith, Ranjan Subramanian, Xiaoqing~Ellen Tan, Binh Tang,
  Ross Taylor, Adina Williams, Jian~Xiang Kuan, Puxin Xu, Zheng Yan, Iliyan
  Zarov, Yuchen Zhang, Angela Fan, Melanie Kambadur, Sharan Narang, Aurelien
  Rodriguez, Robert Stojnic, Sergey Edunov, and Thomas Scialom. 2023.
\newblock \href {https://arxiv.org/abs/2307.09288} {Llama 2: Open foundation
  and fine-tuned chat models}.
\newblock \emph{Preprint}, arXiv:2307.09288.

\bibitem[{{Tringa Sylaj}(2024)}]{Sylaj-MA-image-gen}
{Tringa Sylaj}. 2024.
\newblock Image generation for accessible communication.
\newblock Master's thesis, {Technical University of Munich}.
\newblock Advised and supervised by Miriam Ansch{\"u}tz and Georg Groh.

\bibitem[{Wang et~al.(2022)Wang, Schneider, Alacam, Chaudhury, and
  Biemann}]{wang-motif-complex-words}
Xintong Wang, Florian Schneider, {\"O}zge Alacam, Prateek Chaudhury, and Chris
  Biemann. 2022.
\newblock \href {https://aclanthology.org/2022.lrec-1.263} {{MOTIF}:
  Contextualized images for complex words to improve human reading}.
\newblock In \emph{Proceedings of the Thirteenth Language Resources and
  Evaluation Conference}, pages 2468--2477, Marseille, France. European
  Language Resources Association.

\end{thebibliography}
\appendix
\section{Appendix}
\begin{table*}[!h]
\centering
\begin{tabular}{llll}
\toprule
\textbf{Model} & \textbf{Prompt Limitation} & \textbf{Flagging Content} & \textbf{Resources Used} \\
\midrule
\textbf{SD1\_4} & 77 tokens & Black image & T4 GPU, 15GB RAM \\
\textbf{SD2\_1\_base} & 77 tokens & Black image & T4 GPU, 15GB RAM \\
\textbf{SD\_3} & 77 tokens & Black image & L4 GPU, 24GB RAM \\
\textbf{W\"urstchen} & 77 tokens & Black image & T4 GPU, 15GB RAM \\
\midrule
\textbf{DALL-E-3} & 380 characters & Not processed + warning & Free via Microsoft \\
\textbf{Midjourney} & None & N/A & \$10/month for $\approx$ 200 images \\
\textbf{Artbreeder} & $\sim$ 129 tokens & N/A & Free with multiple accounts \\
\bottomrule
\end{tabular}%
\caption{Comparison of the different models we investigated: Limitations, content flagging, and resource usage}
\label{tab:models_limit}
\end{table*}

\newcommand{\figurecell}[2]{\makecell{#1\\ \centering \includegraphics[height=1.75cm, keepaspectratio]{#2}} }

\begin{table*}[htb]
    \centering
 \begin{adjustbox}{angle=90}
    \begin{tabular}{lm{2.5cm}m{2.5cm}m{2.5cm}m{2.5cm}m{2.5cm}m{2.5cm}m{2.5cm}} 
    \toprule
        \textbf{Model} 
        & \textbf{Multi-family House} 
        & \textbf{Bedroom} 
        & \textbf{Vitamins} 
        & \textbf{Poor Memory Performance}
        & \textbf{Hygiene Products} 
        & \textbf{Yoga} 
        & \textbf{Cafeteria}\\ 
        \midrule
        \textbf{SD\_3} 
        & \figurecell{3}{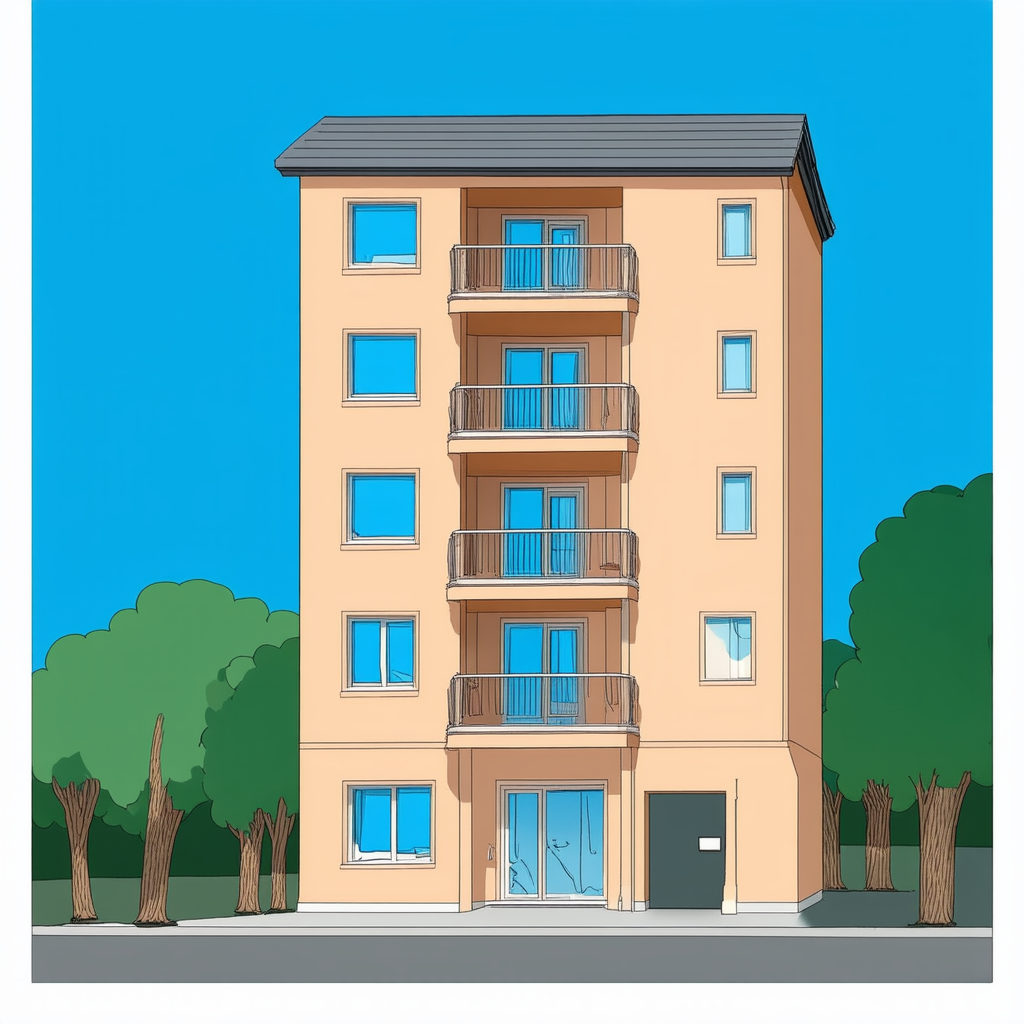}
        & \figurecell{4}{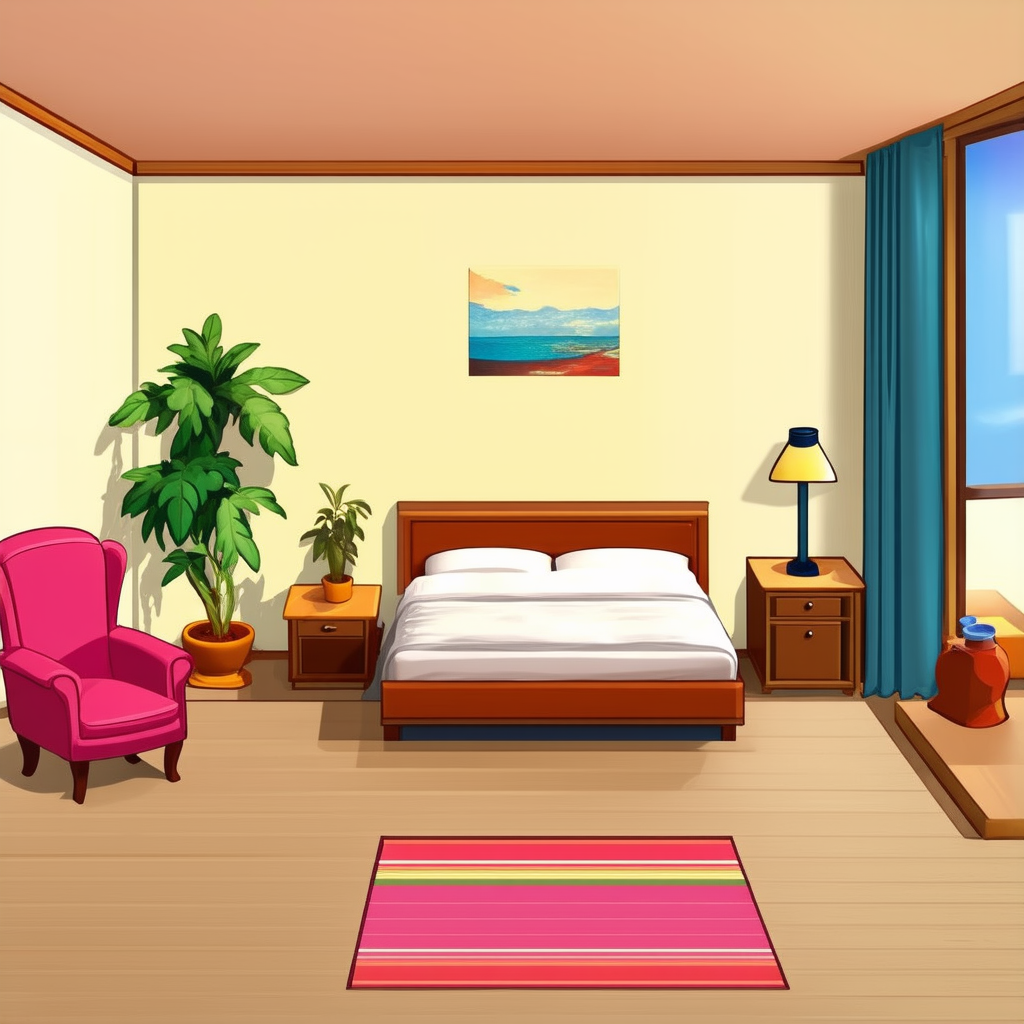}
        & \figurecell{4}{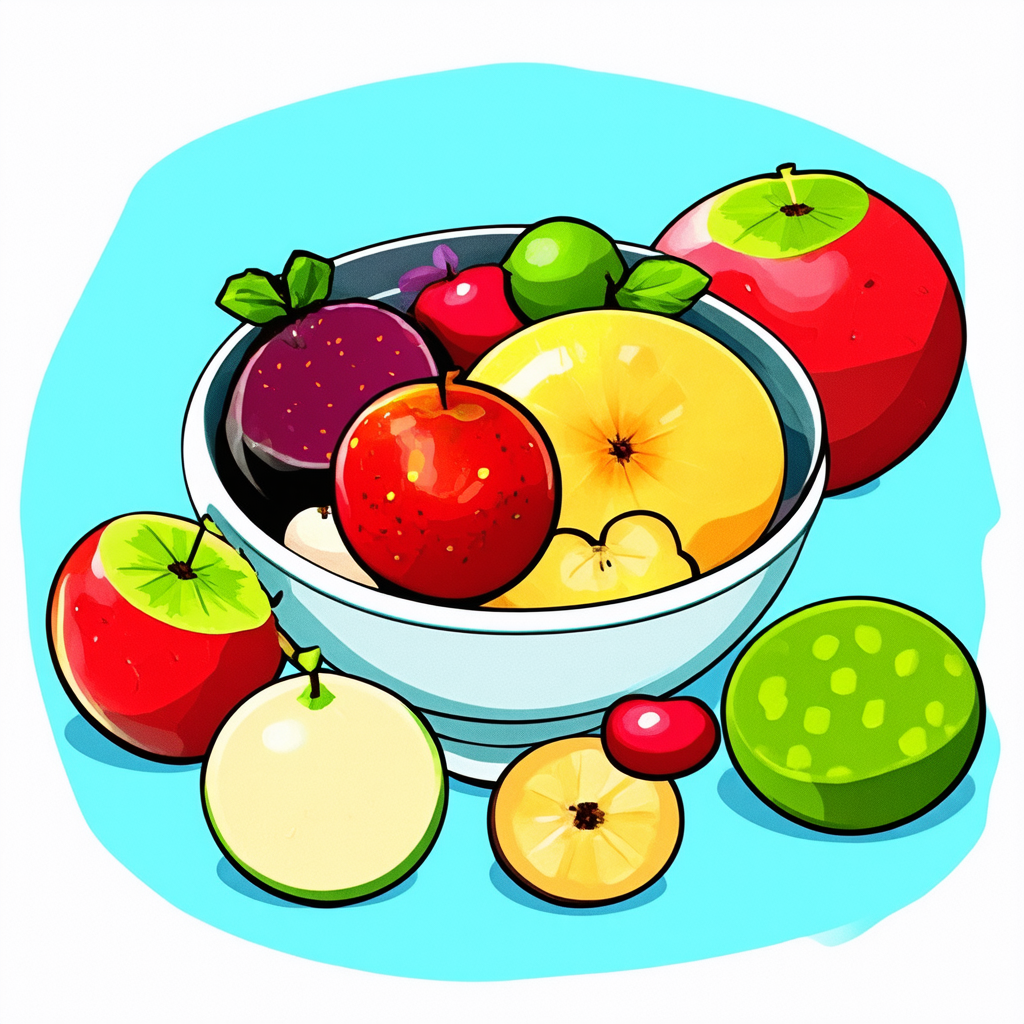} 
        & \figurecell{\textbf{6}}{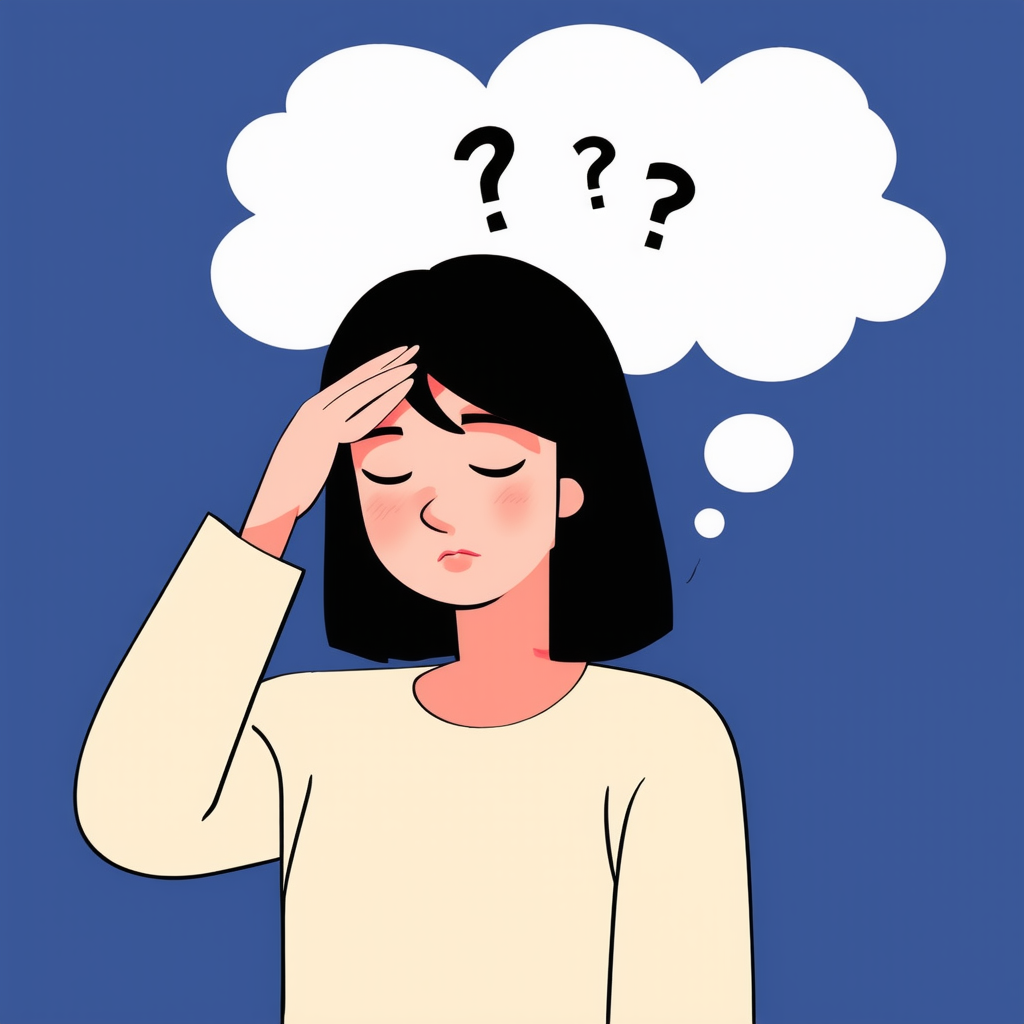} 
        & \centering - 
        & \figurecell{1}{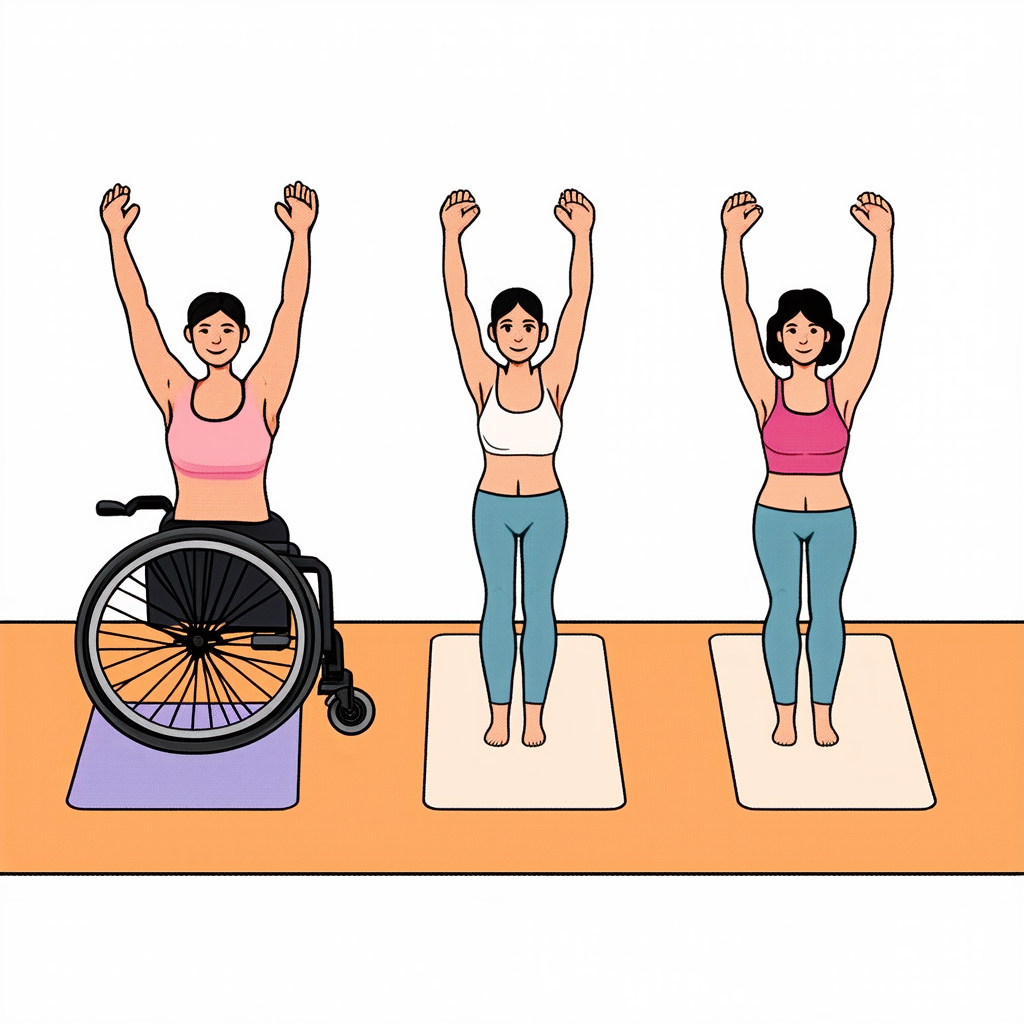} 
        & \makecell{\centering -} \\
         \cdashline{1-8}
         
        \textbf{Wuerstchen} 
        & \figurecell{2}{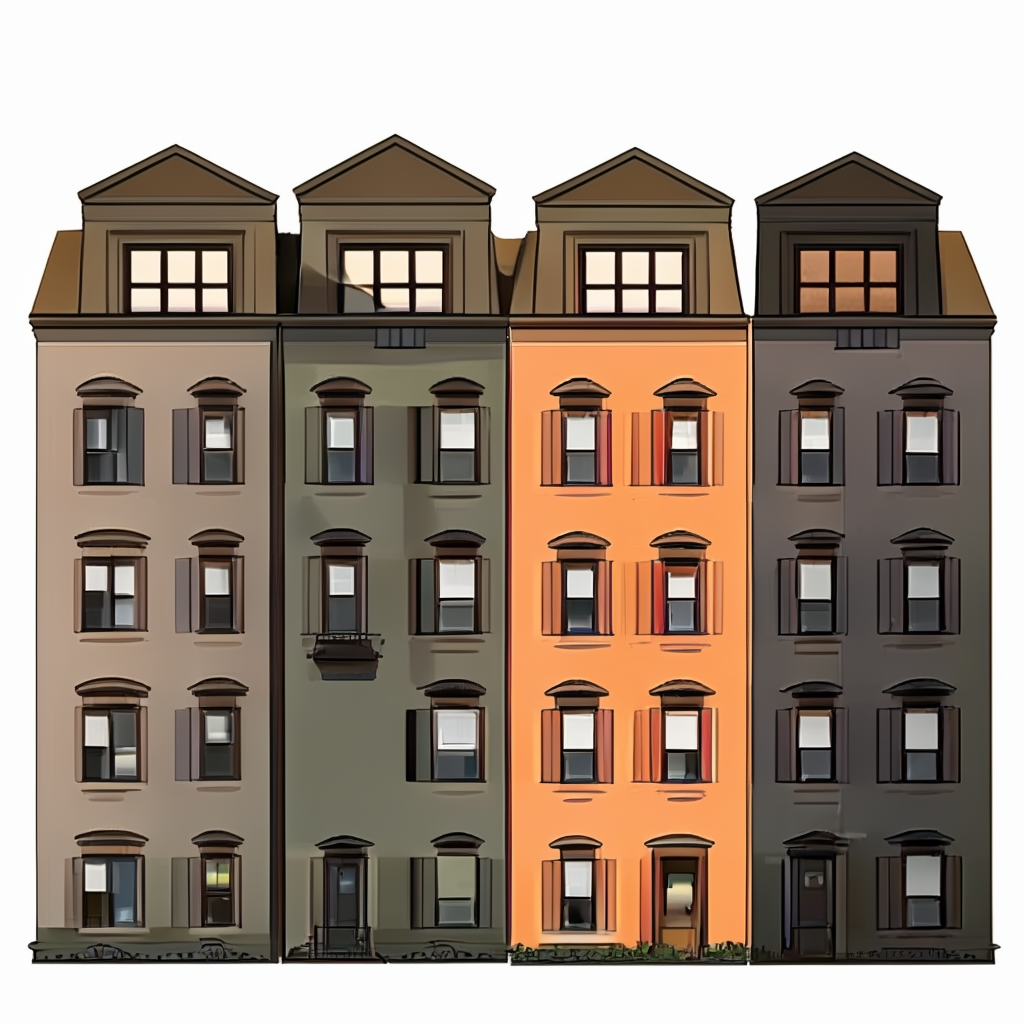} 
        & \centering - 
        & \figurecell{3}{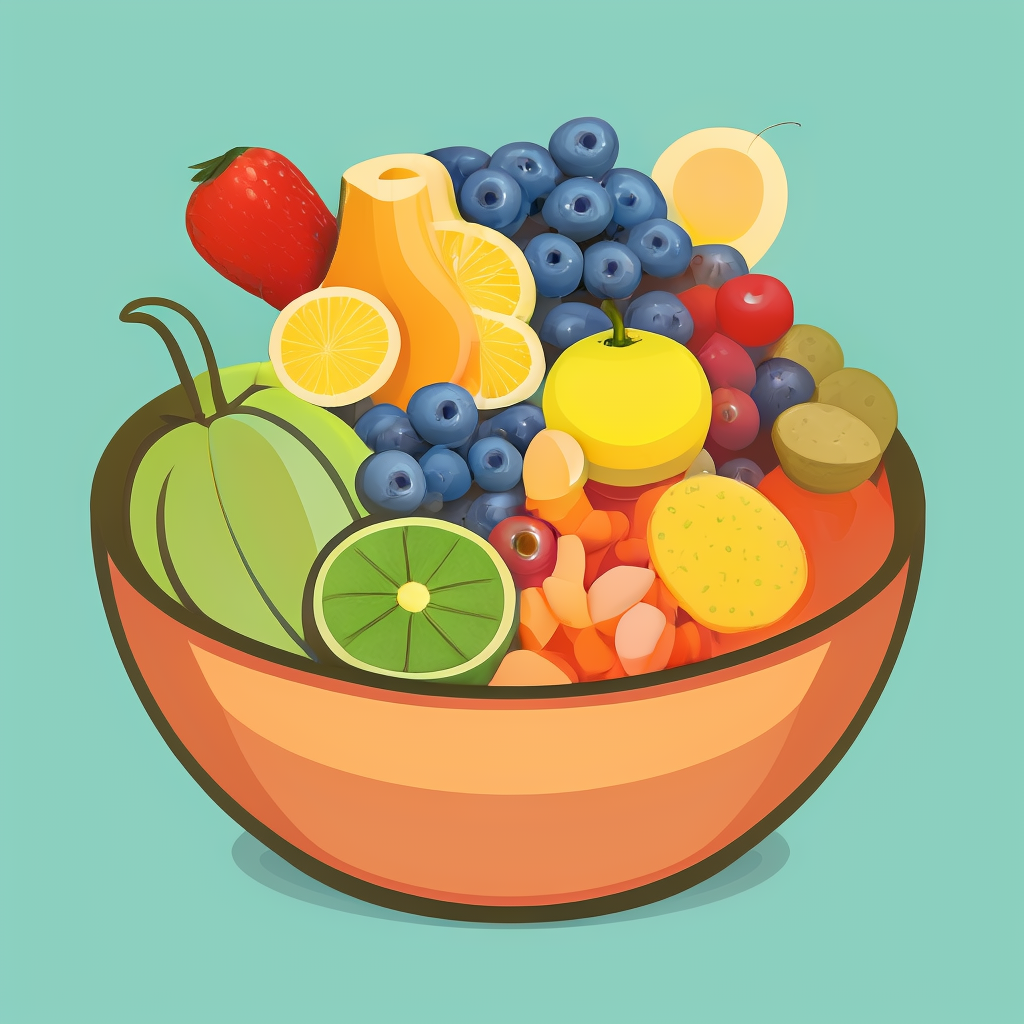}
        & \makecell{\centering -} 
        & \makecell{\centering -} 
        & \makecell{\centering -} 
        & \makecell{\centering -} \\
        \midrule
        
        \textbf{DALLE-3} 
        & \figurecell{\textbf{6}}{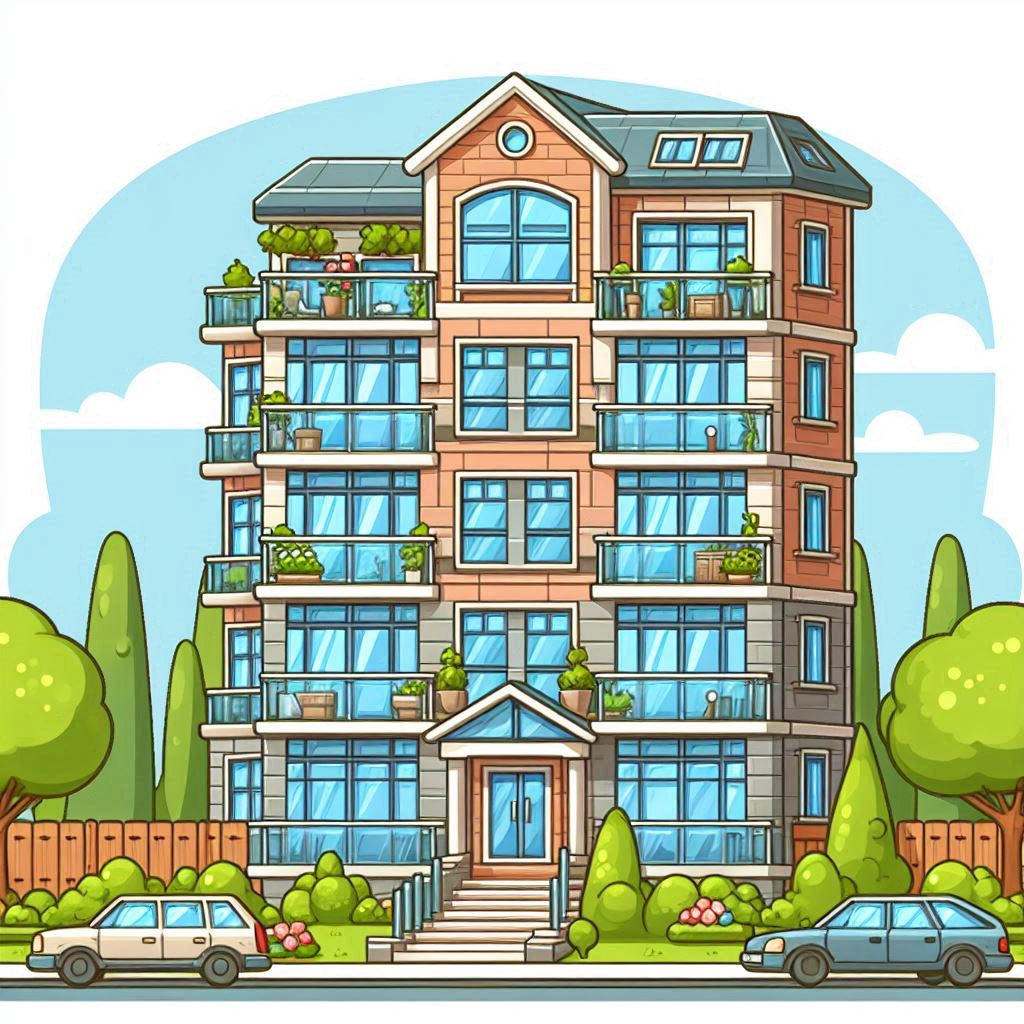} 
        & \figurecell{\textbf{6}}{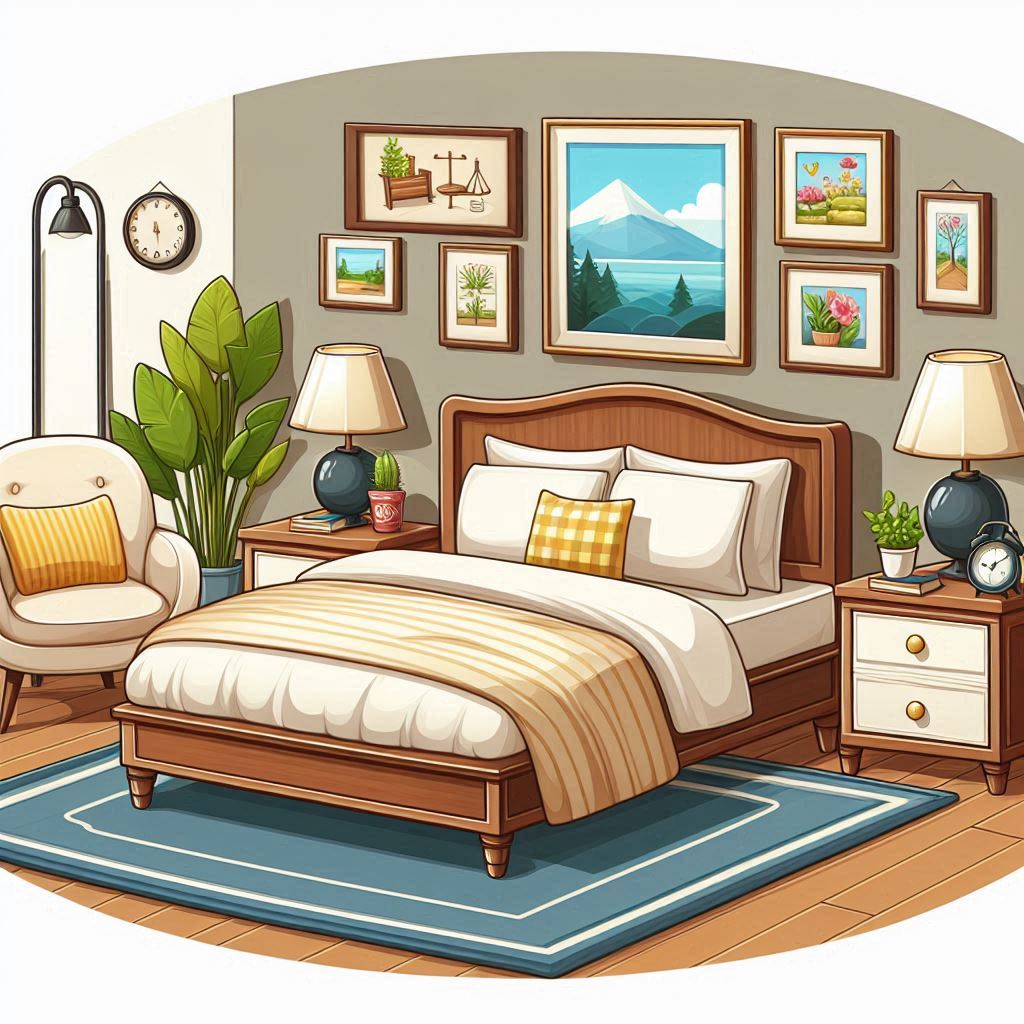} 
        & \figurecell{4}{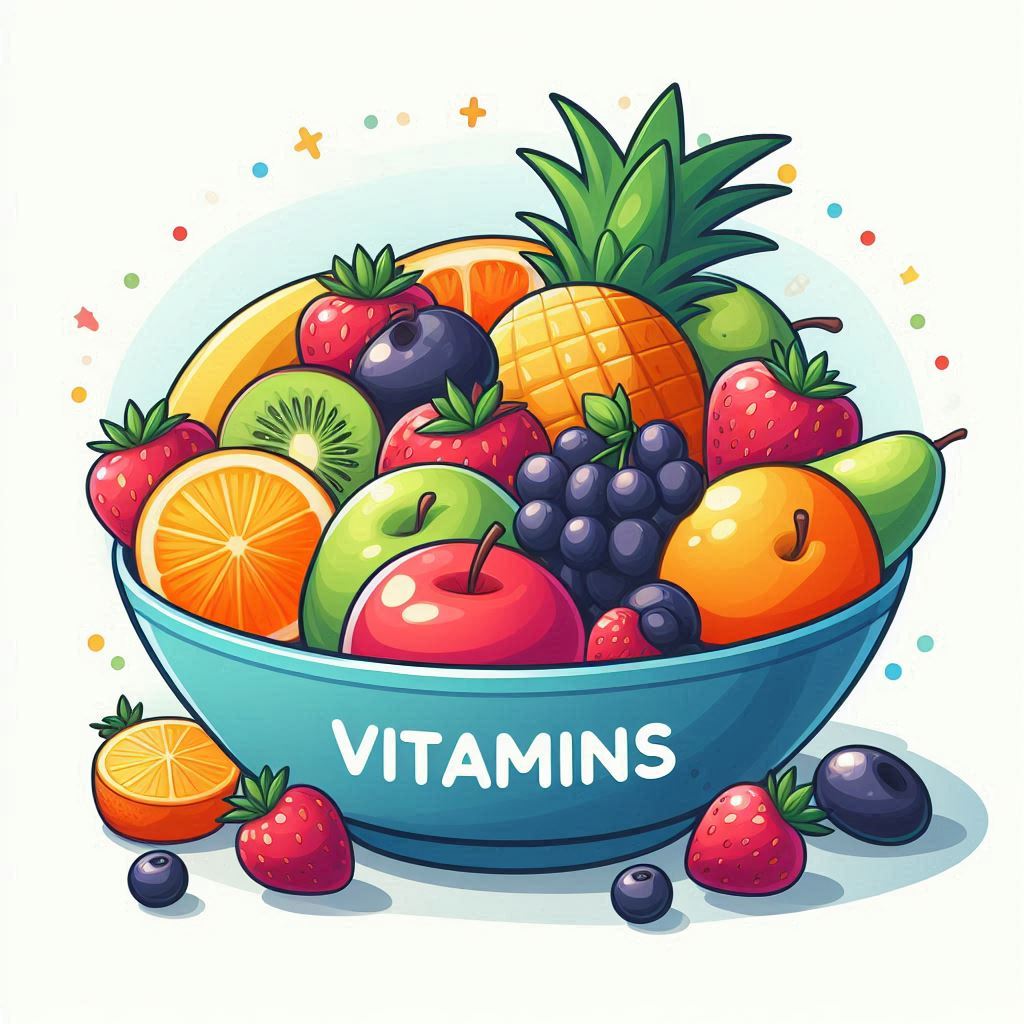} 
        & \figurecell{\textbf{6}}{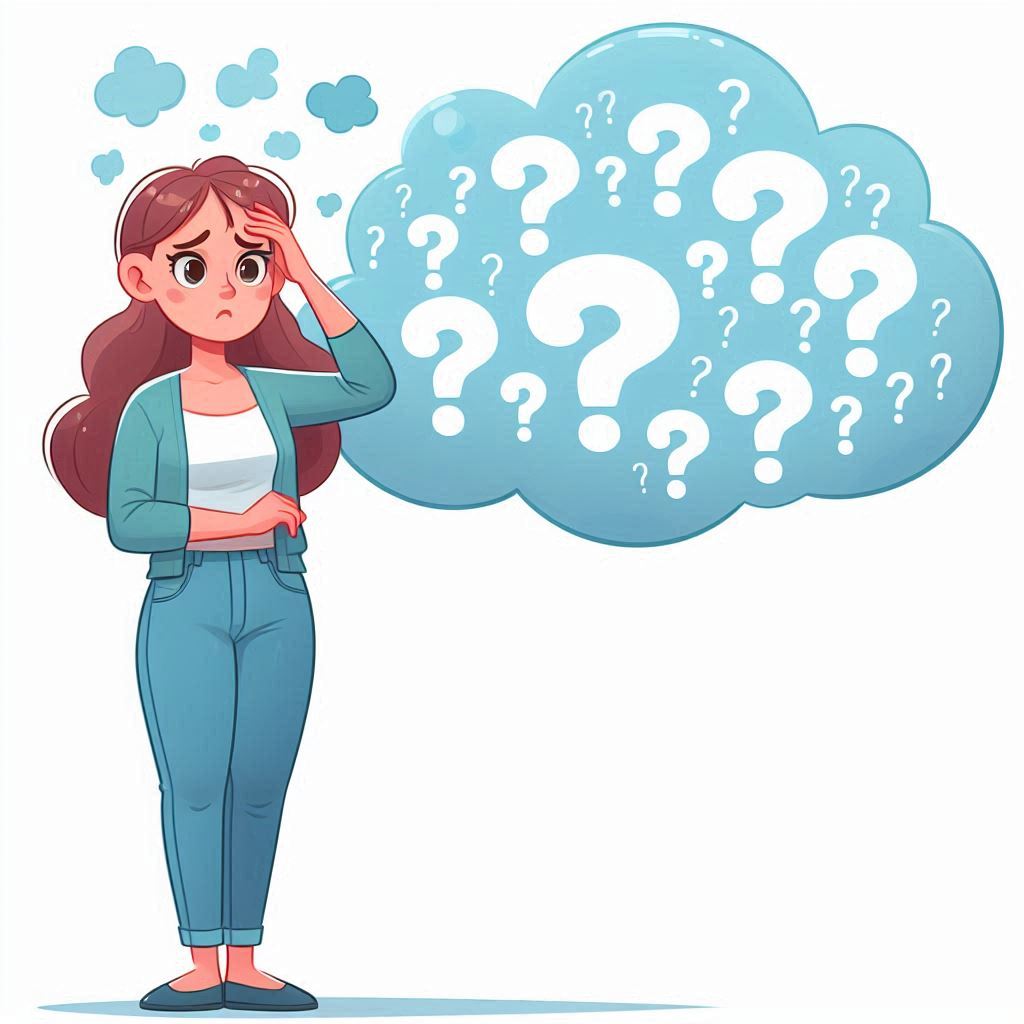}
        & \figurecell{4}{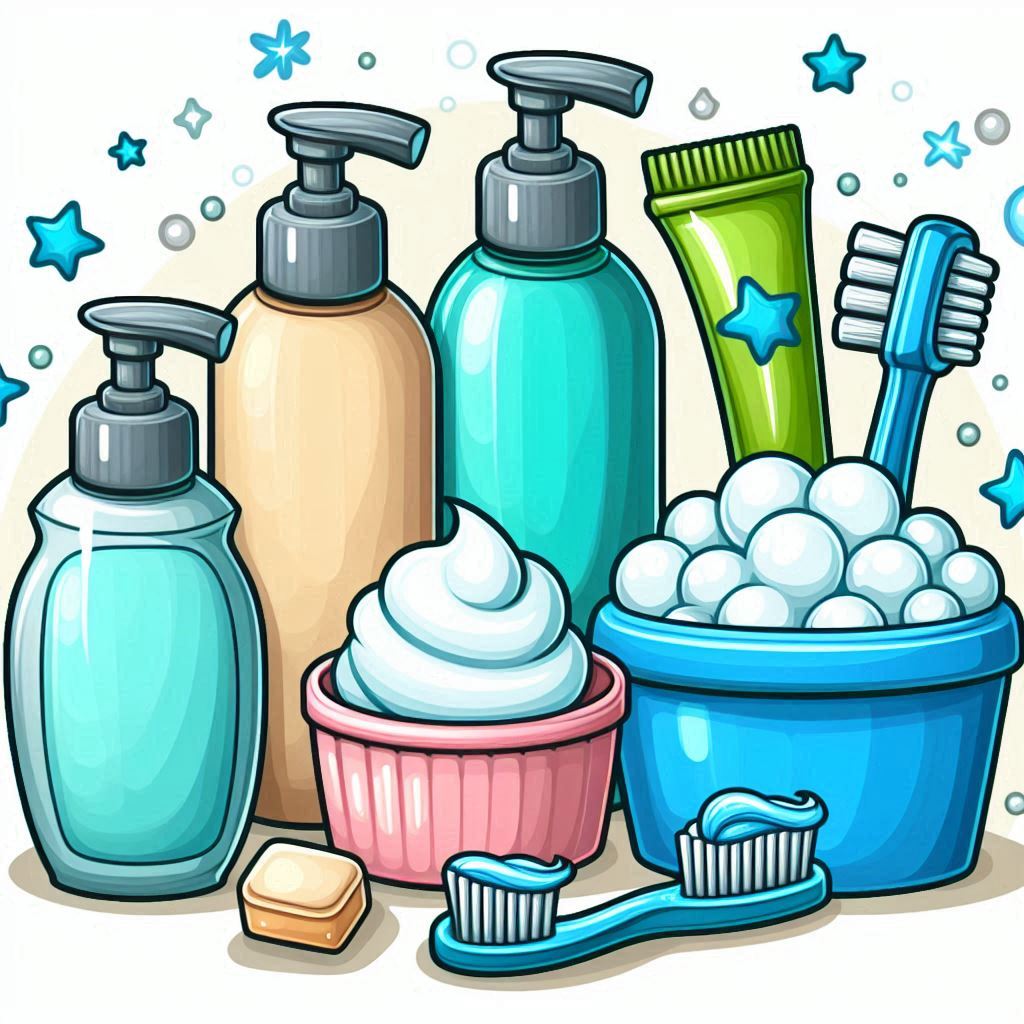}
        & \figurecell{\textbf{2}}{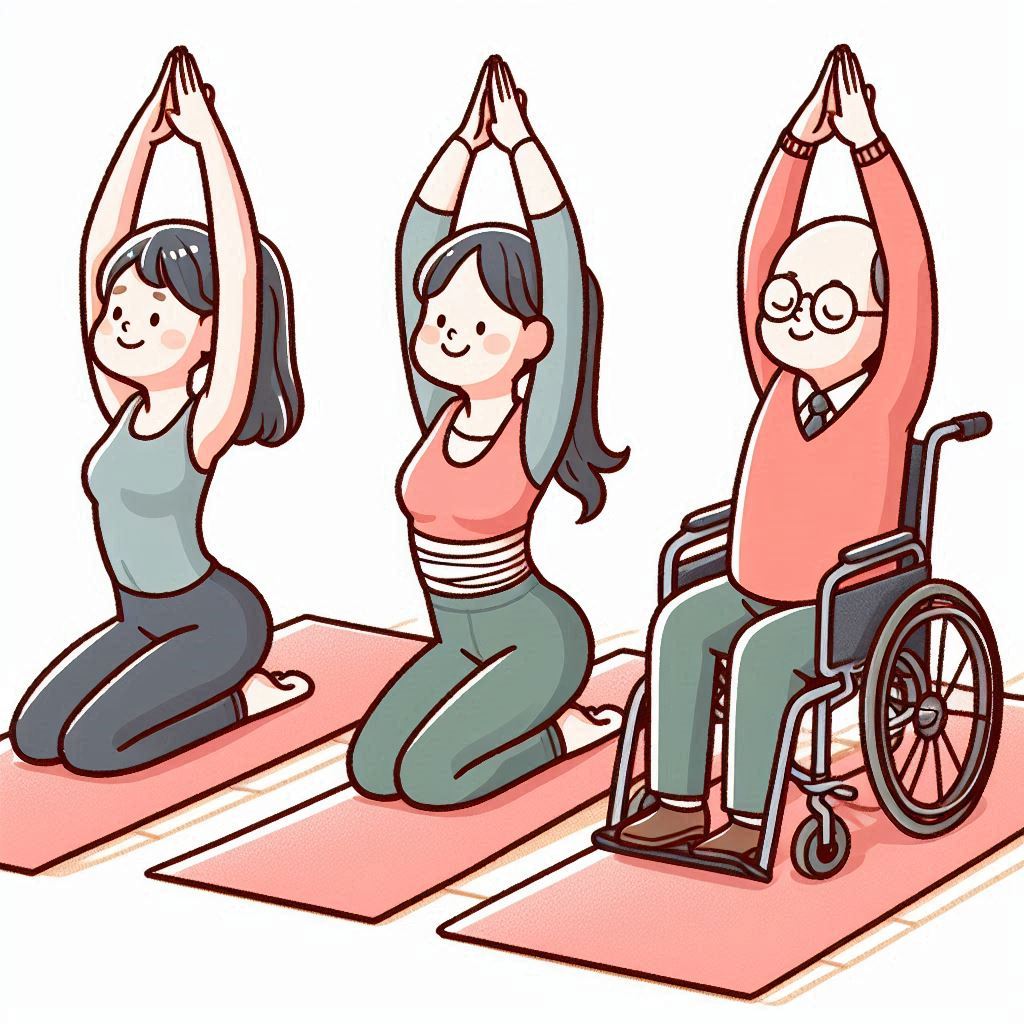}
        & \figurecell{\textbf{4}}{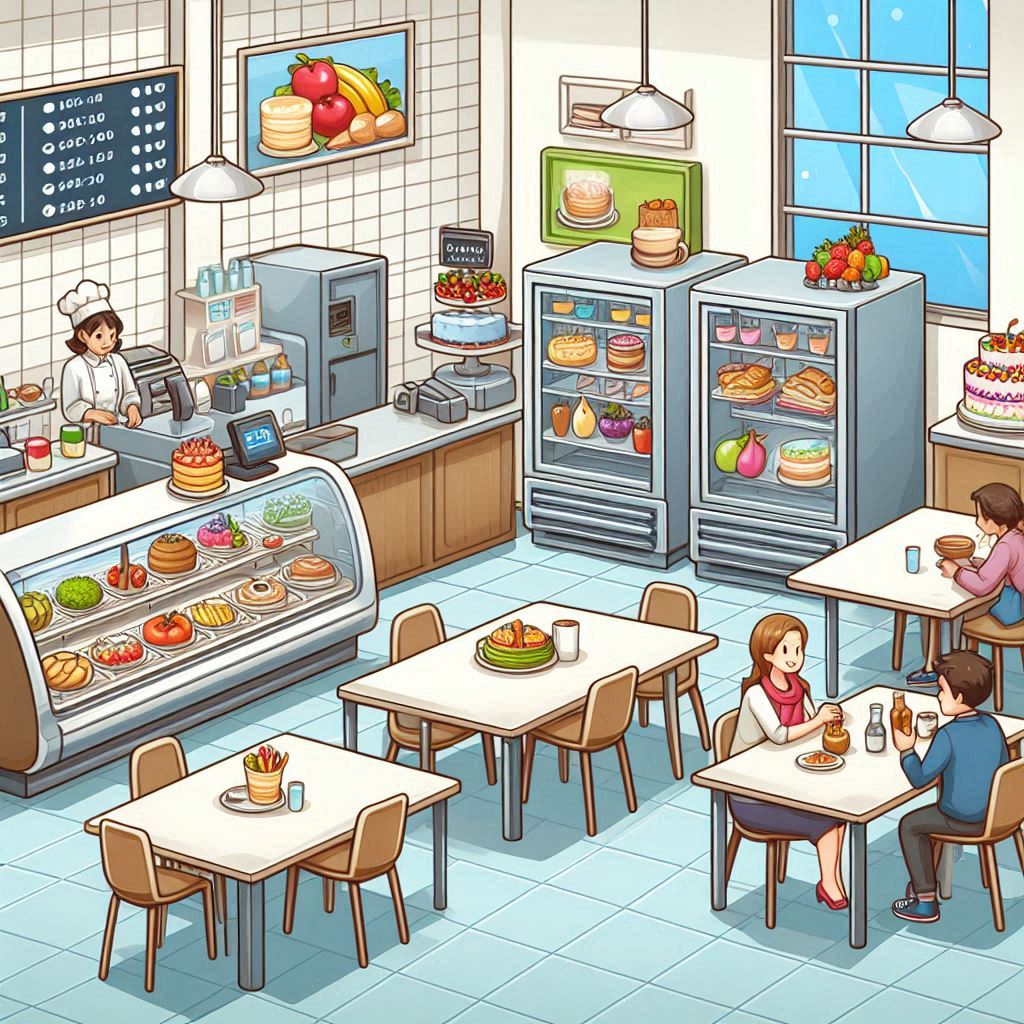}  \\
         \cdashline{1-8}
        
        \textbf{Midjourney} 
        & \figurecell{4}{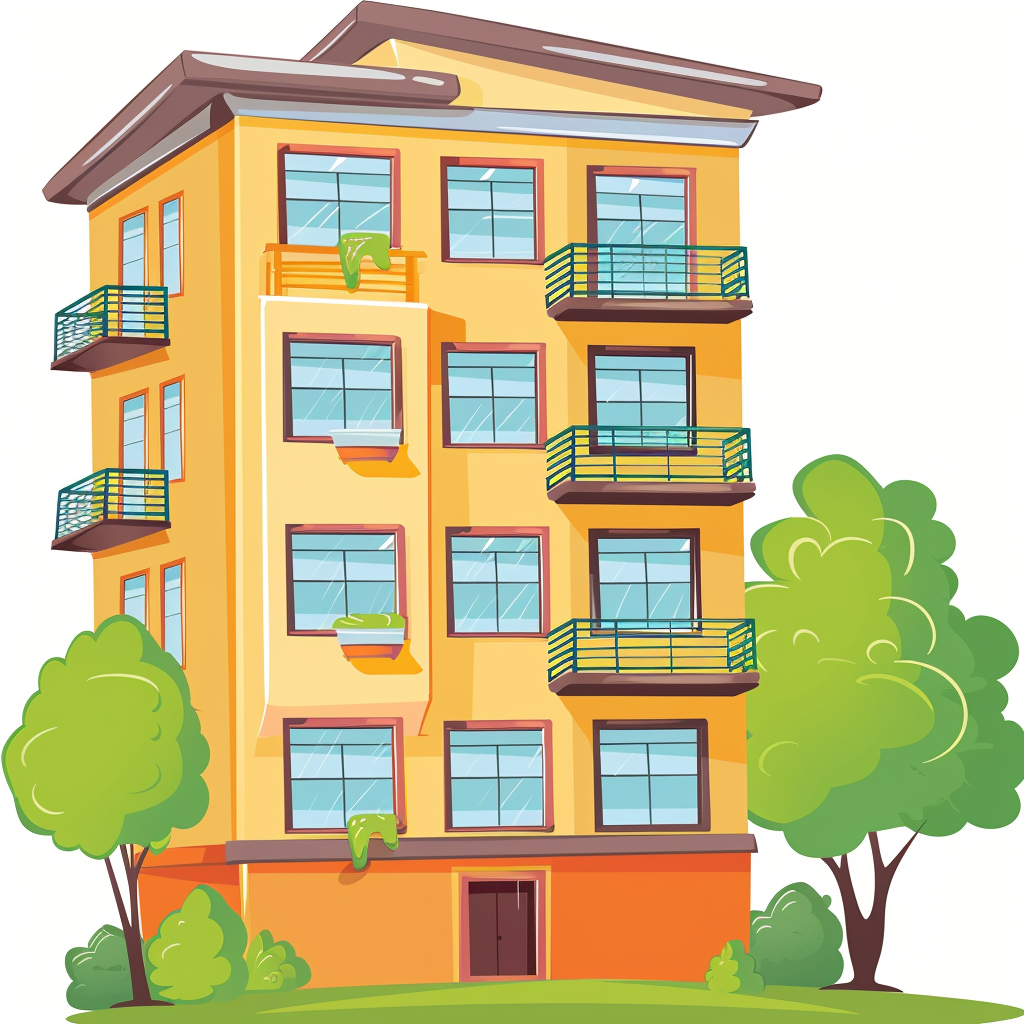} 
        & \figurecell{5}{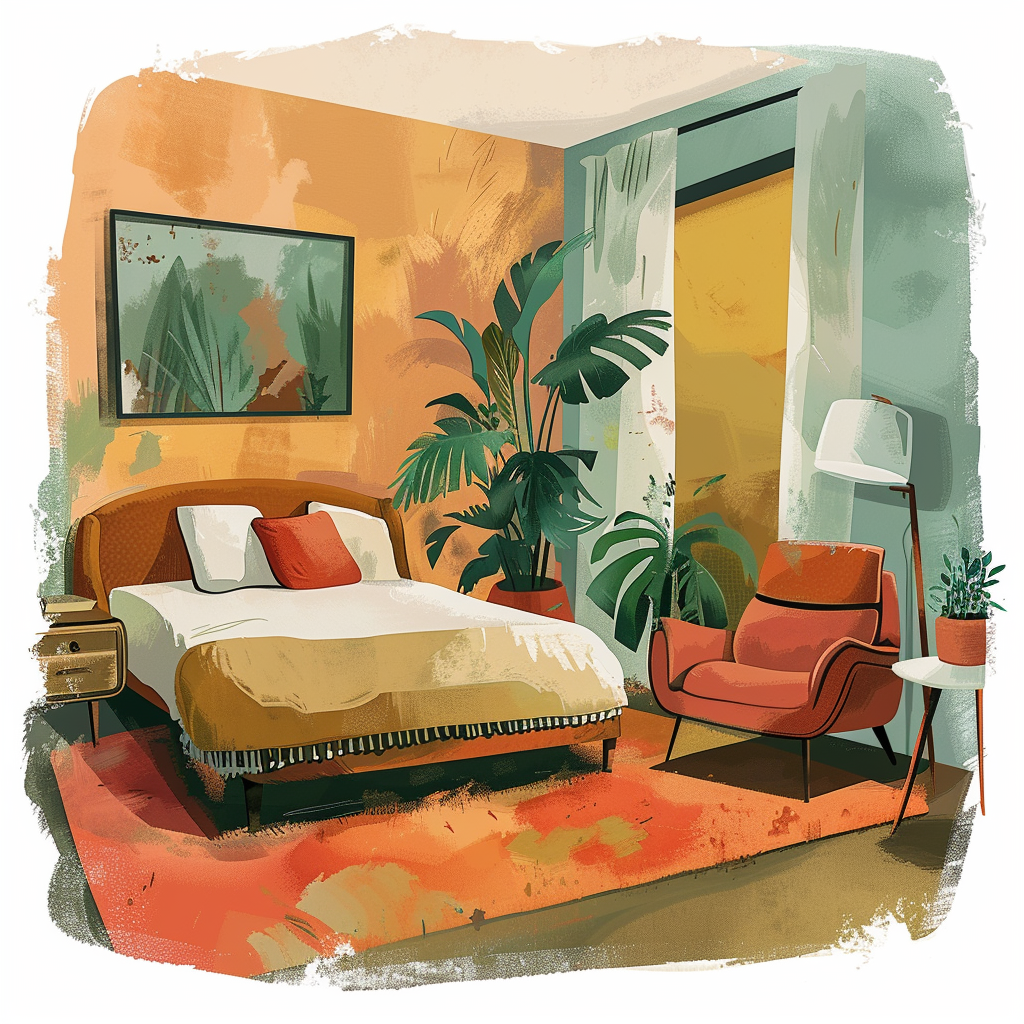} 
        & \figurecell{6}{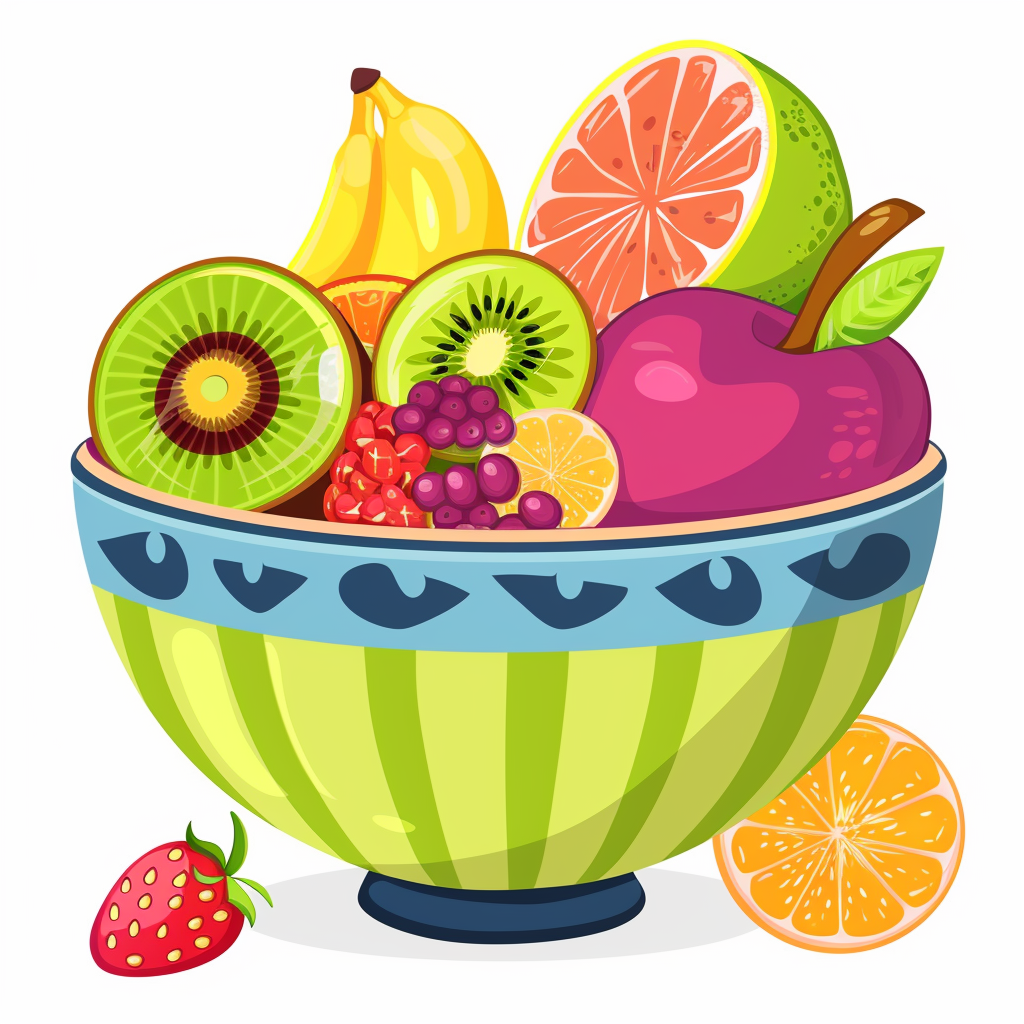}  
        & \figurecell{\textbf{6}}{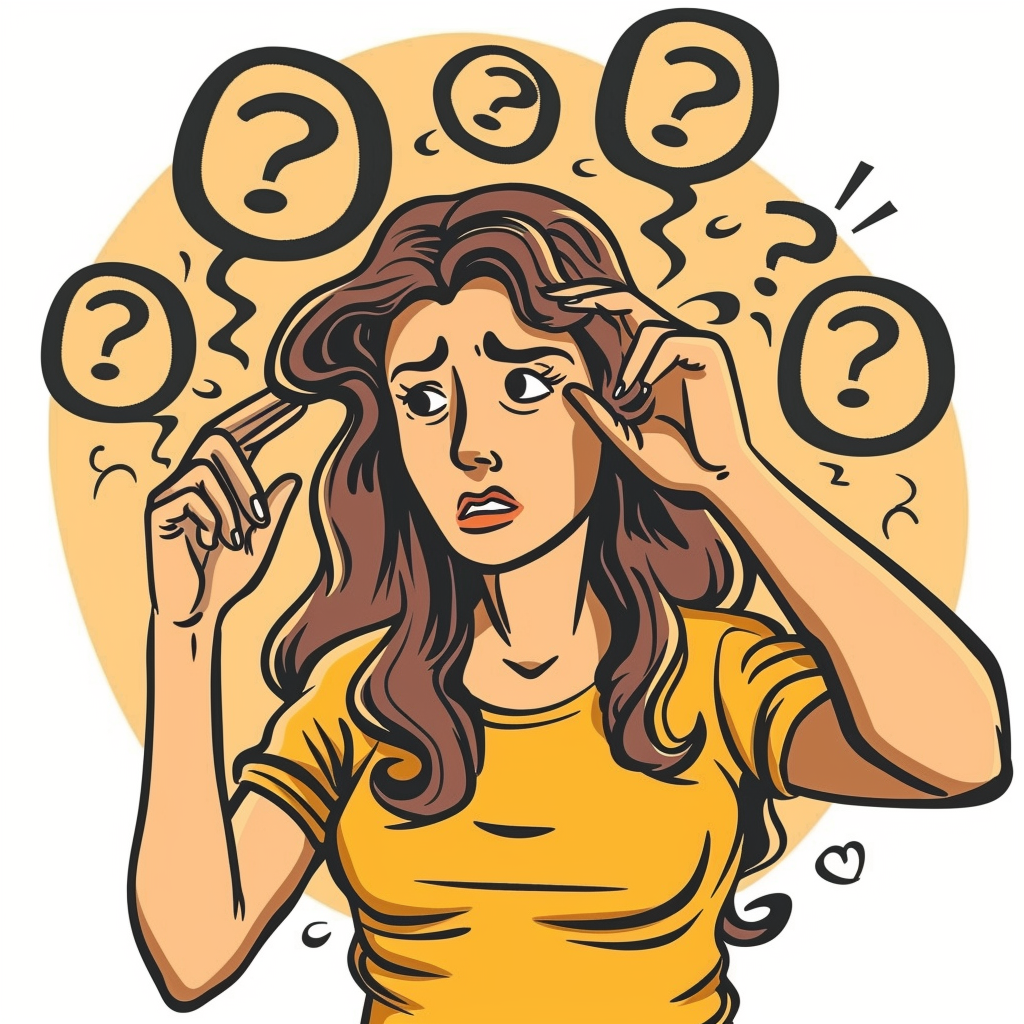}
        & \figurecell{\textbf{5}}{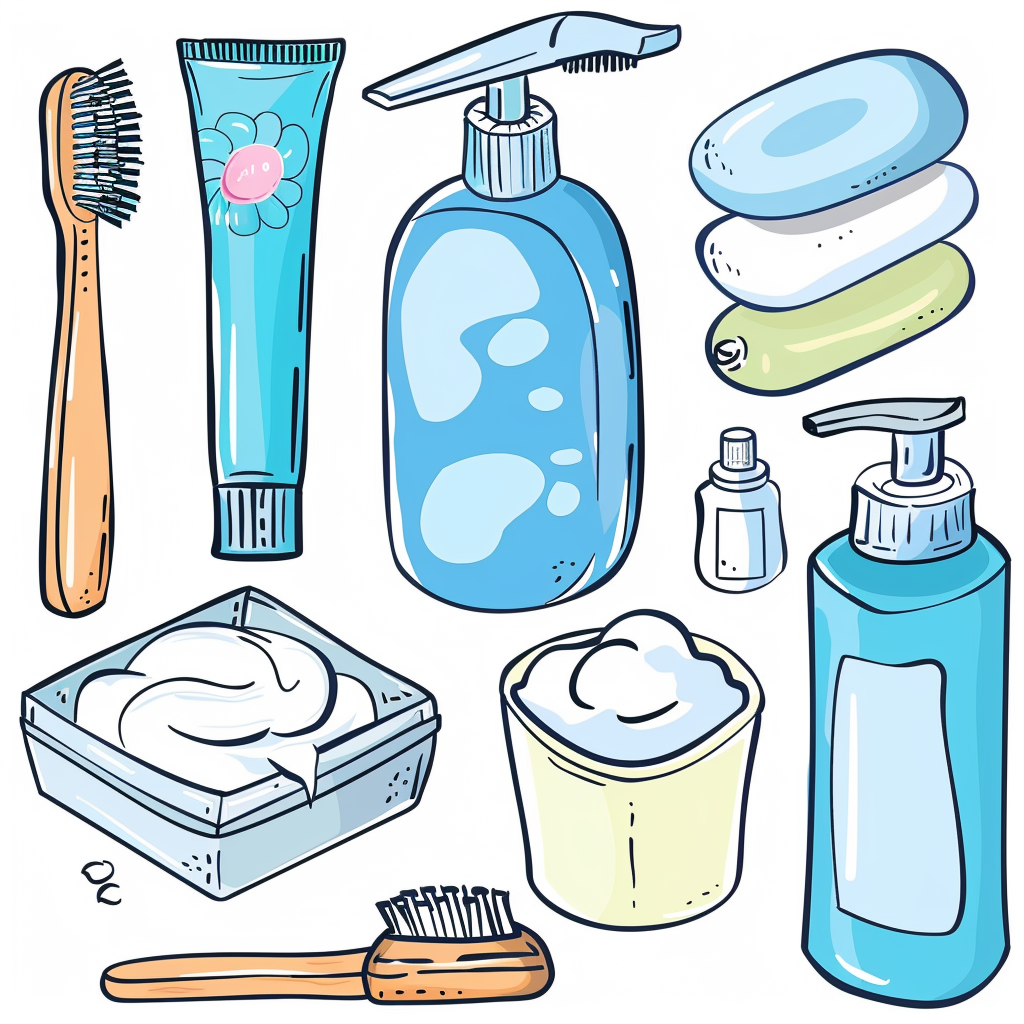} 
        & \figurecell{\textbf{2}}{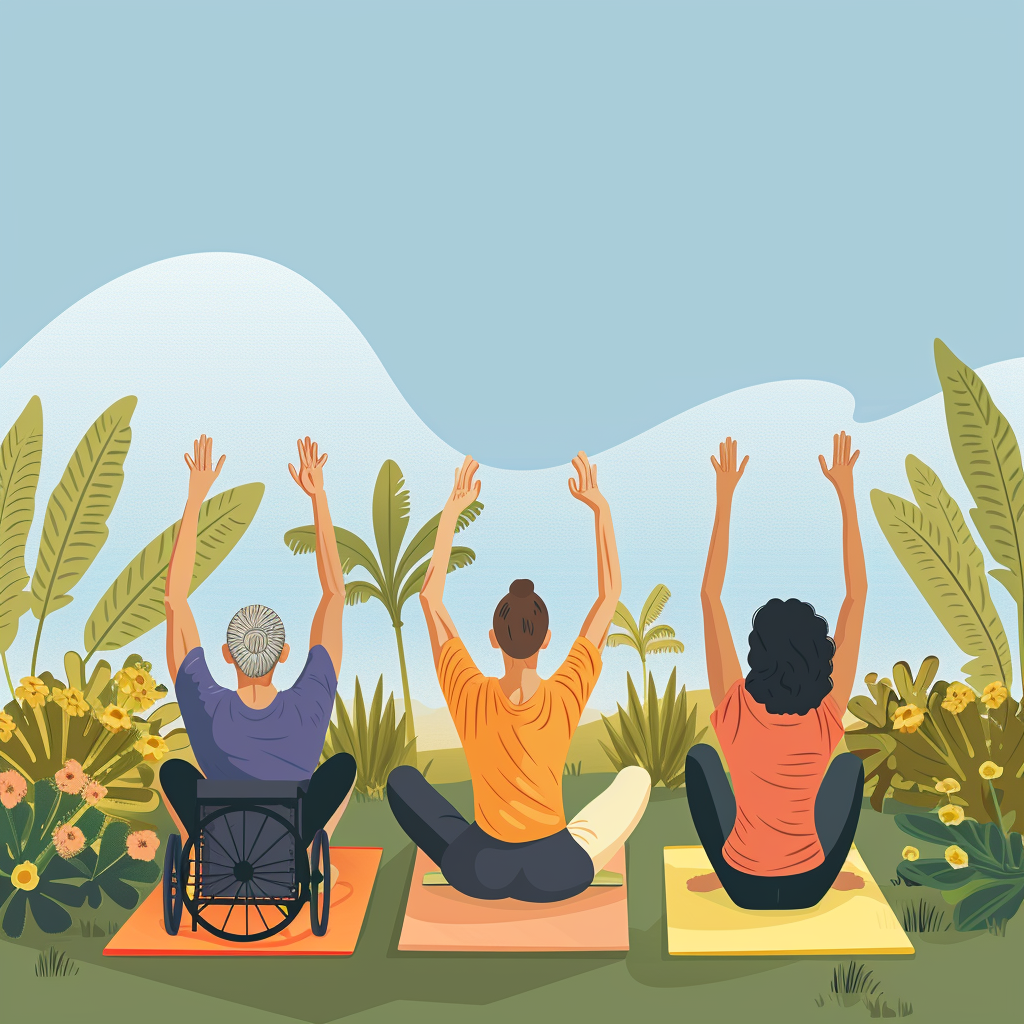}  
        & \figurecell{\textbf{4}}{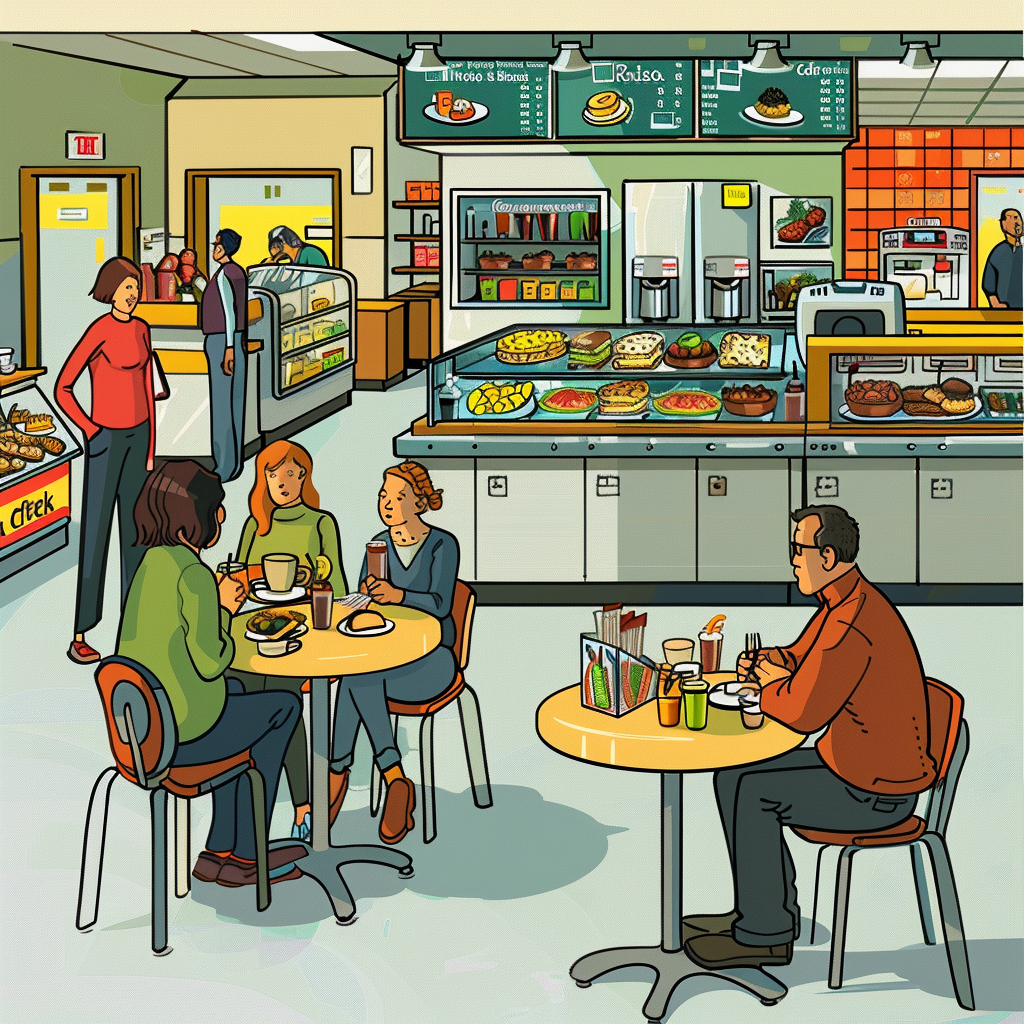} \\
         \cdashline{1-8}
        
        \textbf{artbreeder} 
        & \figurecell{5}{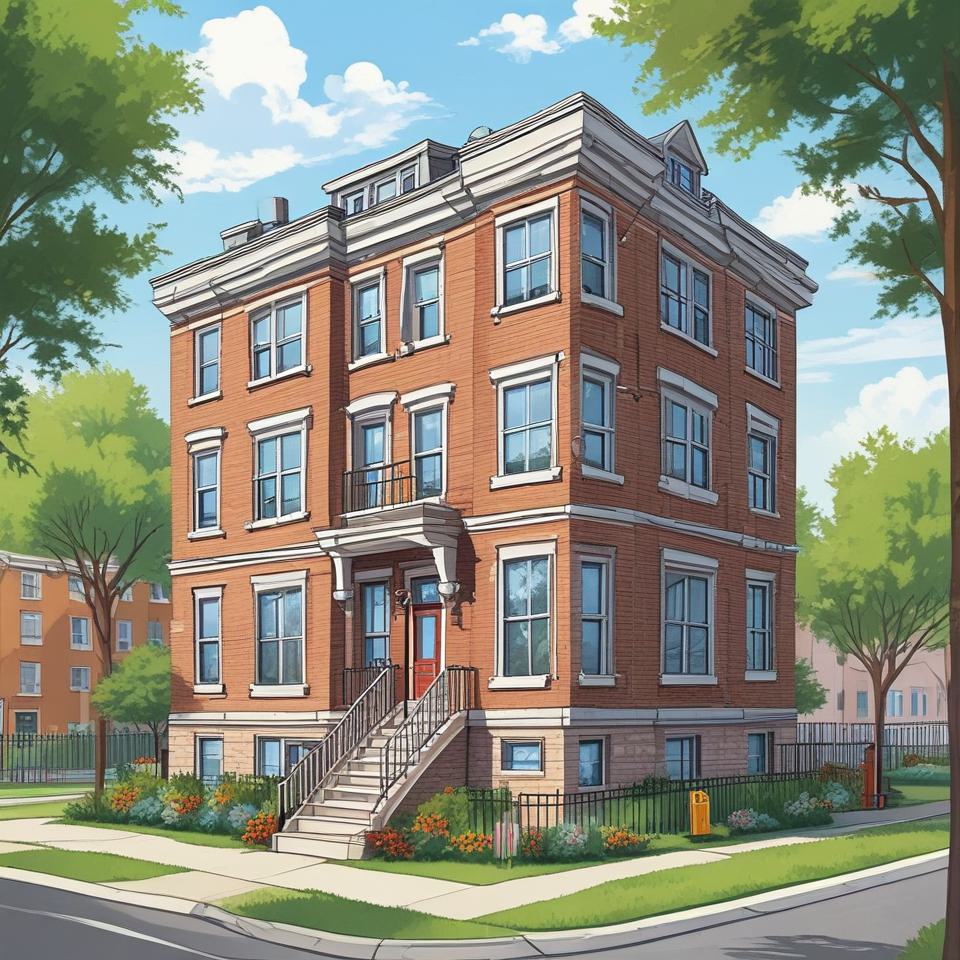}
        & \figurecell{1}{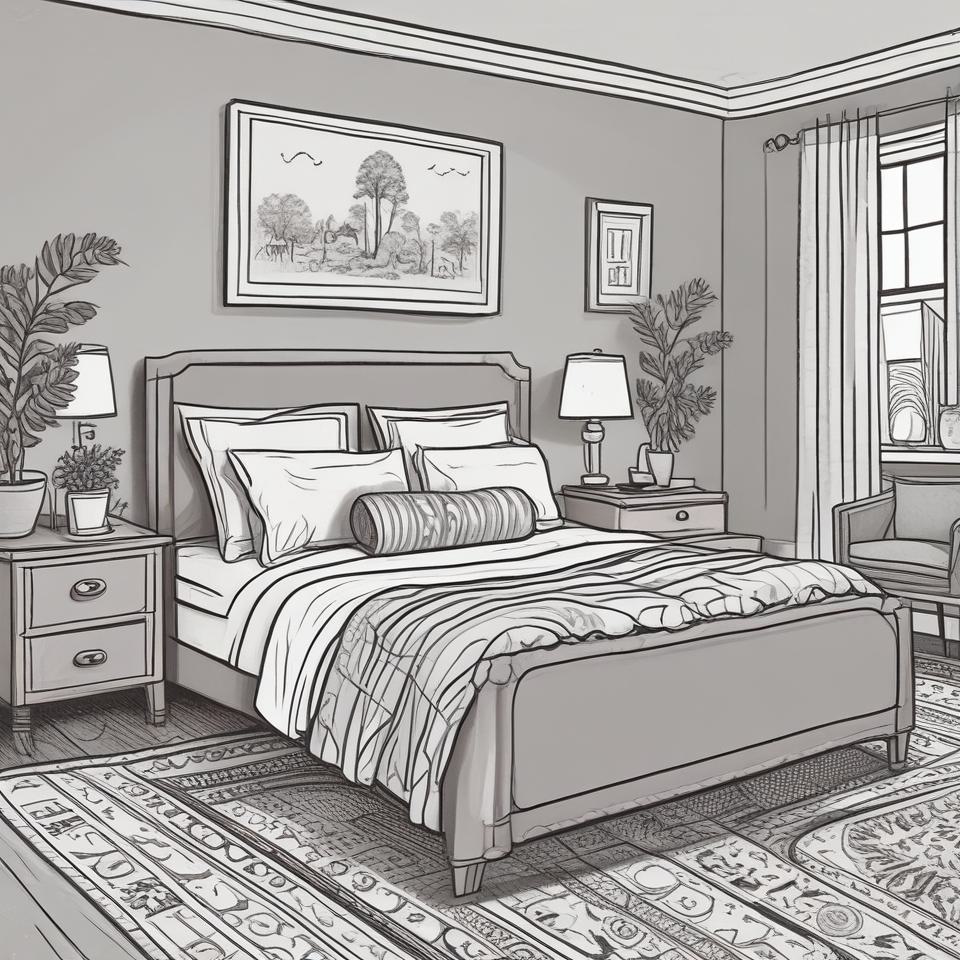}
        & \figurecell{3}{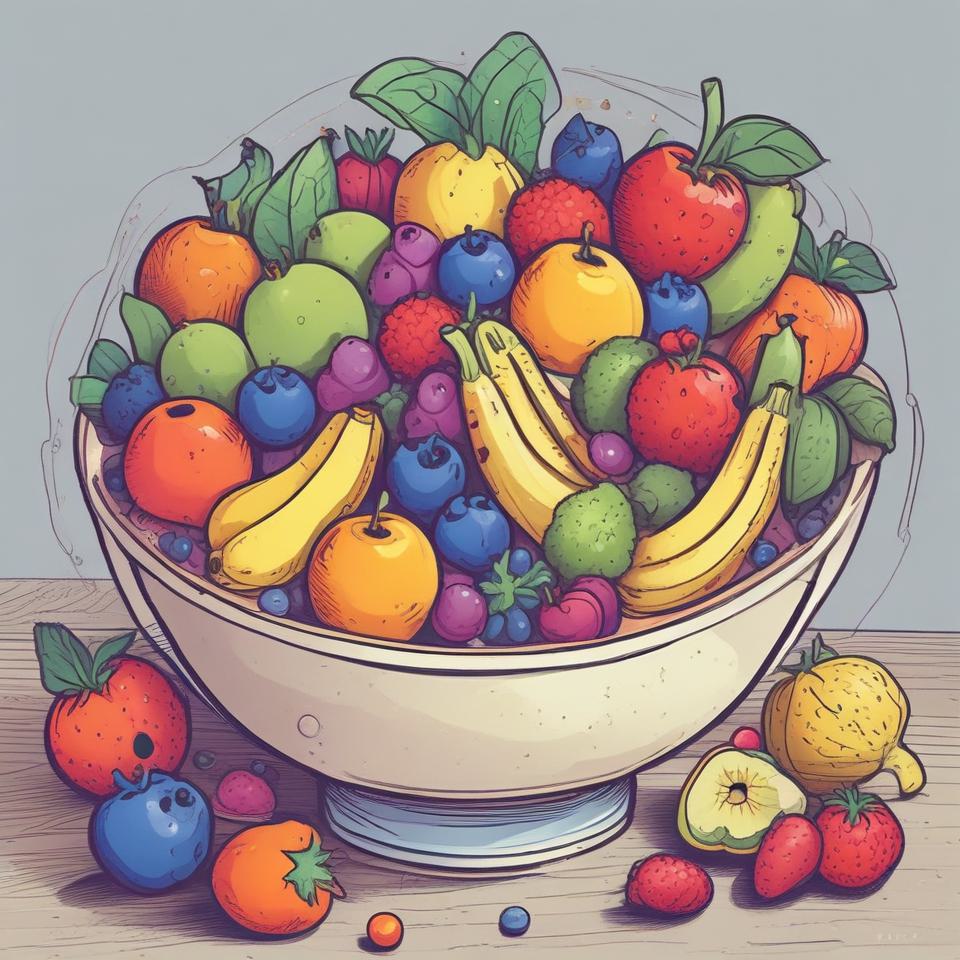} 
        & \makecell{\centering -}
        & \figurecell{4}{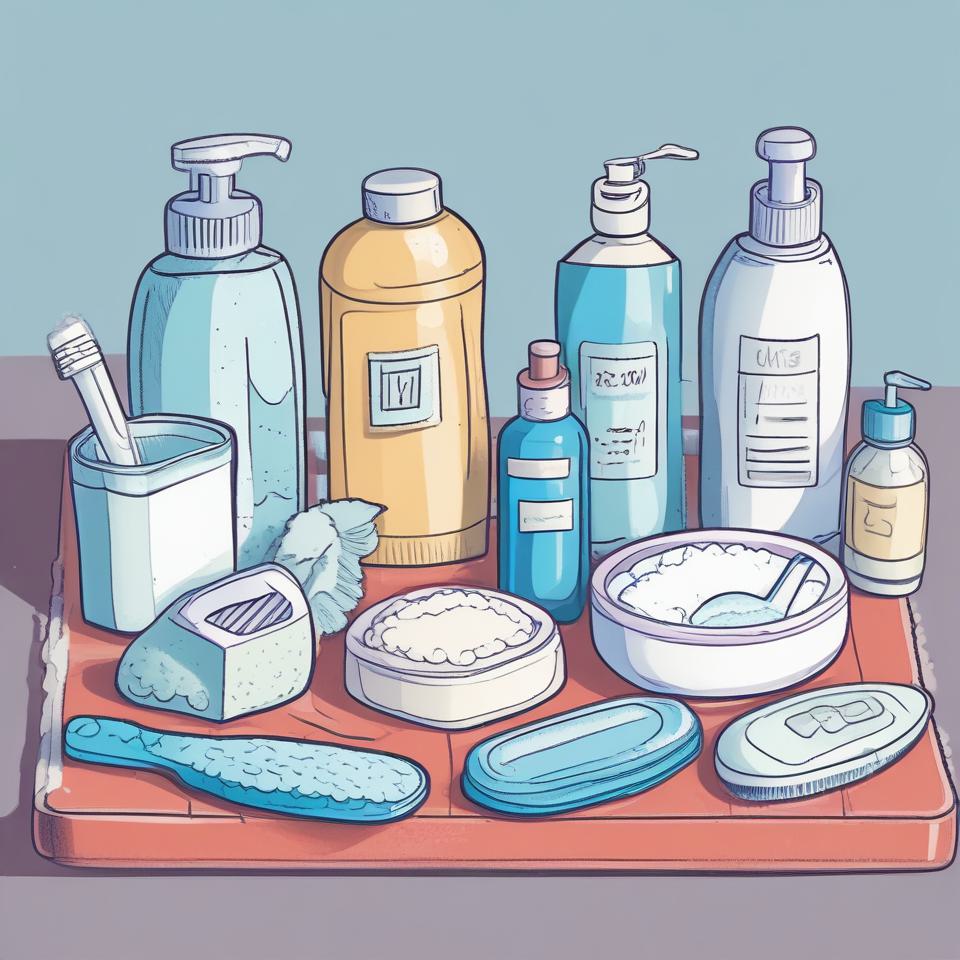}
        & \makecell{\centering -}
        & \makecell{\centering -} \\ 
        \midrule
        
        \textbf{References} 
        & \figurecell{4}{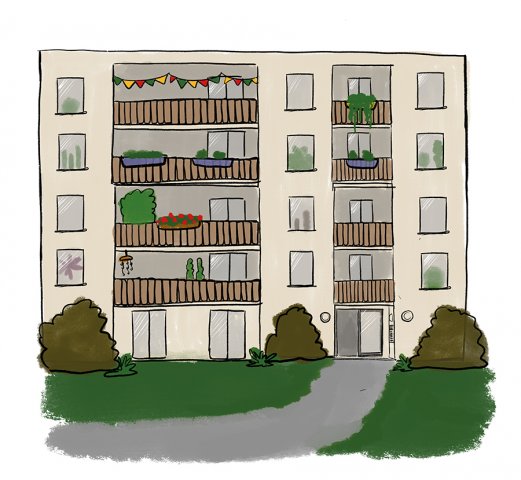} 
        & \figurecell{3}{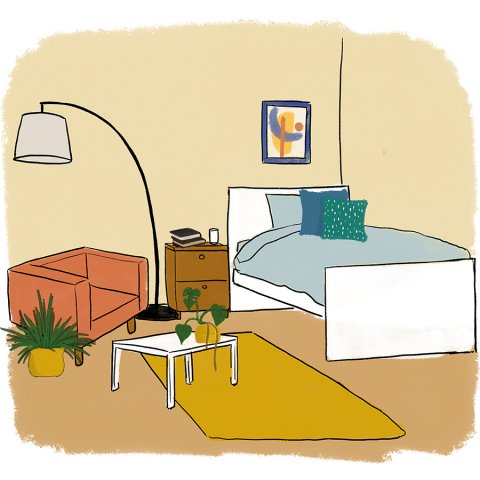}
        & \figurecell{\textbf{7}}{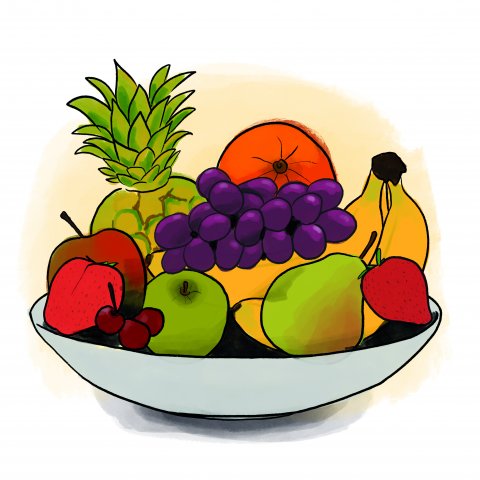} 
        & \figurecell{5}{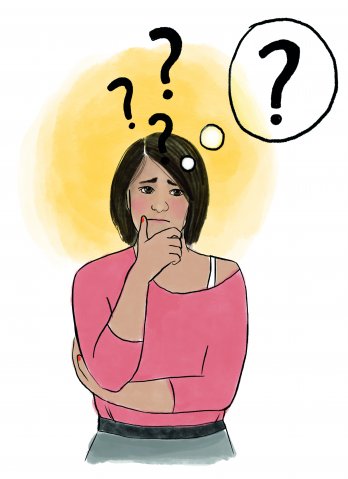} 
        & \figurecell{4}{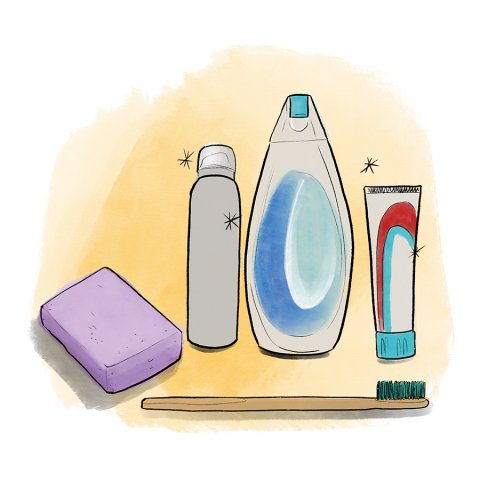} 
        & \figurecell{\textbf{2}}{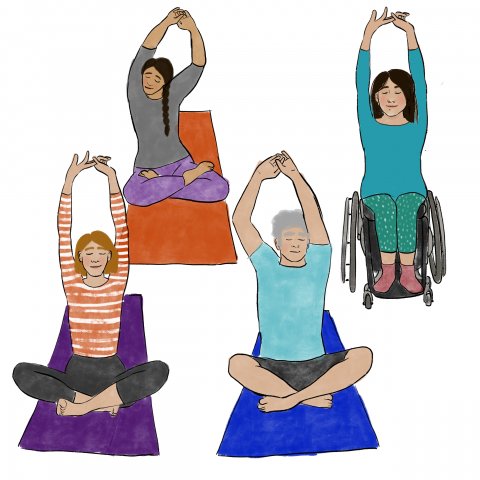} 
        & \figurecell{2}{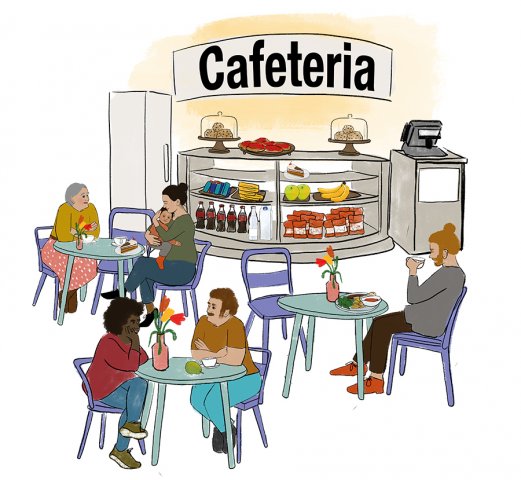} \\
        \bottomrule
    \end{tabular}
 \end{adjustbox}
    \caption{Number of votes from the target group during their review session. We only included images that the \ls{} expert deemed suitable for the target group. Thus, no images from Stable Diffusion 1 and 2 were shown.}
    \label{tab:target_group_votes}
\end{table*}

\end{document}